%% file: iclr2026_conference.tex
\documentclass{article} 
\usepackage{iclr2026_conference,times}
\input{math_commands.tex}

\usepackage[hyphens]{url} %
\usepackage{hyperref}
\usepackage{url}

\usepackage{enumitem}
\usepackage{multirow}
\usepackage{multicol}
\usepackage{graphicx}
\usepackage{arydshln}
\usepackage{algorithm2e}
\usepackage{tcolorbox}
\usepackage[table]{xcolor}
\usepackage{booktabs}      

\usepackage[utf8]{inputenc}
\usepackage[T1]{fontenc}

\title{OViP: Online Vision-Language Preference Learning For VLM Hallucination}

\author{Shujun Liu\\Fudan University\And Siyuan Wang\\University of Southern California\And Zejun Li\\Fudan University\AND Jianxiang Wang\\ByteDance\And Cheng Zeng\\ByteDance\And Zhongyu Wei\\Fudan University}

\newcommand{\ovip}{OViP}

\arxivcopy
\begin{document}

\maketitle
\let\oldthefootnote\thefootnote
\renewcommand{\thefootnote}{\fnsymbol{footnote}}
\footnotetext{Please contact us at \texttt{shujuanliu24@m.fudan.edu.cn}.}
\let\thefootnote\oldthefootnote

\begin{abstract}
  Large vision-language models (LVLMs) remain vulnerable to hallucination, often generating content misaligned with visual inputs. 
  Although recent training-based approaches aim to mitigate hallucination, they typically rely on predefined or randomly edited negative samples that do not reflect actual model errors, thus limiting training efficacy.
  In this work, we propose an Online Vision-language Preference Learning (\ovip) framework that dynamically constructs contrastive training data based on the model’s own hallucinated outputs. By identifying semantic differences between sampled response pairs and synthesizing negative images using a diffusion model, \ovip{} generates more relevant supervision signals in real time. This failure-driven training enables adaptive alignment of both textual and visual preferences. Moreover, we refine existing evaluation protocols to better capture the trade-off between hallucination suppression and expressiveness. Experiments on hallucination and general benchmarks demonstrate that \ovip{} not only reduces hallucinations while preserving core multi-modal capabilities, but also substantially improves training efficiency. Code is available at \url{https://github.com/lsjlsj35/Online-Vision-Language-Preference-Learning-for-VLM-Hallucination}.
\end{abstract}

\section{Introduction}
\input{sections/introduction}

\section{Methodology}
\input{sections/method}

\section{Experiment}
\input{sections/experiment}
\section{Further Study}
\input{sections/further_study}

\section{Related Work}
\input{sections/related_work.tex}
\section{Conclusion}
\input{sections/conclusion}
\newpage
\section*{Ethics Statement}
\input{appendix/Ethic_statement}

\section*{Reproducibility Statement}
We provide detailed descriptions of the training and evaluation setups in Appendix~\ref{app.evaluation} and Appendix~\ref{app.ablation}. In addition, we include anonymized training and evaluation code, instructions for running the experiments, and information on accessing the relevant datasets in the supplementary materials.

\bibliographystyle{iclr2026_conference}
\bibliography{refs}

\appendix
\newpage
\section{Evaluation}\label{app.evaluation}
\input{appendix/evaluation}

\section{Experiments}\label{app.ablation}
\input{appendix/ablation}
\section{Algorithm}
The pseudocode is at \autoref{tab:app:algorithm}.

\section{Efficiency and Time Consuming}\label{app.eff}

\input{appendix/efficiency}
\section{Limitations}
\input{appendix/limitations}

\ifarxiv\else
\section{Use of LLMs}
We used existing large language models solely for language polishing and minor coding assistance. The models were not involved in the design of experiments, development of research ideas, or analysis of results.
\fi
\newpage
\section{Prompts for Judgment and Negative Image Generation}
\input{appendix/prompt}

\end{document}

%% file: math_commands.tex
\usepackage{amsmath,amsfonts,bm}

\def\eqref#1{equation~\ref{#1}}

\def\1{\bm{1}}

\DeclareMathAlphabet{\mathsfit}{\encodingdefault}{\sfdefault}{m}{sl}
\SetMathAlphabet{\mathsfit}{bold}{\encodingdefault}{\sfdefault}{bx}{n}



%% file: sections/introduction.tex
Large vision-language models (LVLMs)~\citep{flamingo,minigptv2,sharegpt4v,llavabenchinthewild, llava15} have demonstrated remarkable performance across a wide range of multi-modal tasks~\citep{instructblip,blip2,qwen,qwen2vl} by integrating pre-trained visual encoders with large language models (LLMs) to process and generate language grounded in visual inputs. However, LVLMs continue to struggle with persistent hallucination issues~\citep{li2023evaluatinghallu,bai2024hallucination}, often exhibiting incorrect references to visual content~\citep{liu2024survey,zhou2023analyzing,bai2024hallucination}. 
These errors manifest as misattributing object properties, describing nonexistent entities, or fabricating spatial relationships that do not align with the image. Such inconsistencies undermine the model's faithfulness to the input and hinder further reasoning capabilities, significantly limiting the reliability of LVLMs in real-world applications.

Recent success of Direct Preference Optimization (DPO)~\citep{dpo} in LLMs alignment has inspired the exploration of multi-modal DPO to mitigate hallucination in LVLMs~\citep{rlhfv,rlaifv,v-dpo,halva}. However, early efforts directly extend the original DPO designs from LLMs to LVLMs by constructing preference pairs solely on textual responses given the same image input, primarily focusing on response-side preference optimization and showing limited effectiveness. 
Recent advancements incorporate additional preference pairs conditioned on varying image inputs while keeping the same response, optimizing both visual and textual preference optimization~\citep{mdpo,svco,chip}. This paradigm provides a complementary training signal that encourages the model to attend more closely to visual content.

However, prior work mainly relies on existing paired datasets~\citep{svco} or expert-defined patterns to construct negative image inputs, using techniques such as random cropping~\citep{mdpo}, noise disruption~\citep{povid}, object removal~\citep{AdaViP2504}, or human/LLMs generated element-replaced response for image editing~\citep{v-dpo}. 
These strategies are typically not explicitly tied to model failures, resulting in distribution misalignment between the generated negatives and the model’s actual hallucination behavior, thereby offering limited improvement and failing to support adaptive and continual online\footnote{We adopt the LLM community’s convention of using ``online'' to denote ``on-policy'' in RL.} learning~\citep{guo2024direct}. To address these limitations, we propose a failure-guided negative generation strategy that directly targets self-generated hallucinated responses, enabling the real-time creation of more in-distribution counterexamples through text-to-image generation. Specifically, we sample and filter positive and negative response pairs from the model’s textual outputs. Then we employ LLMs to generate an image prompt based on the negative response, particular emphasizing its differences from the positive response, and subsequently synthesize the corresponding negative image using a diffusion model~\citep{zhang2023text}.

\begin{figure}[t]
\centering
\includegraphics[width=\linewidth]{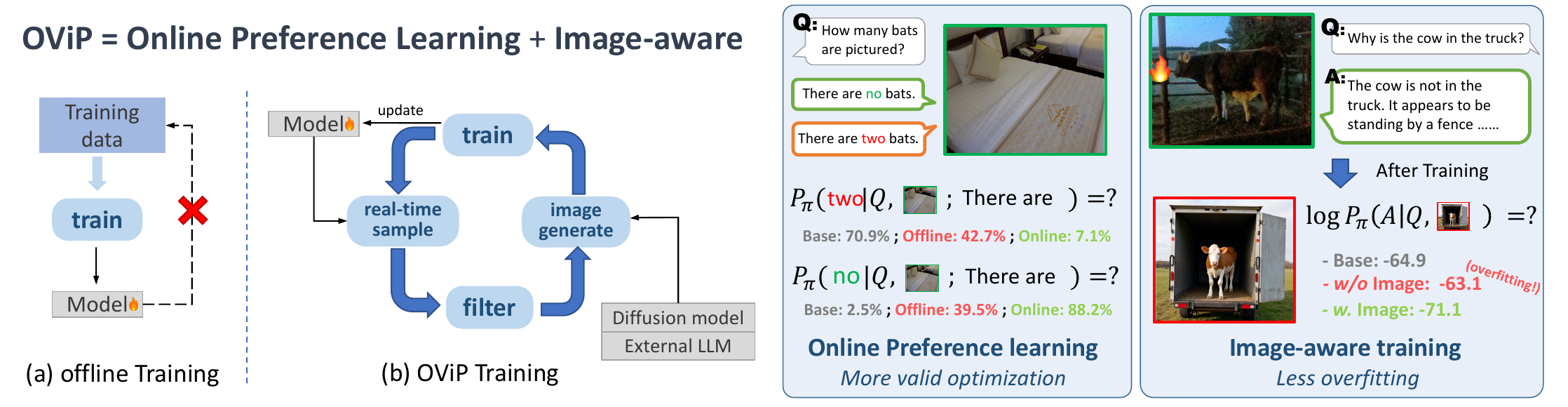}
\caption{
Offline training (a) relies on static, predefined datasets and fails to adapt to the model's evolving failure patterns, limiting its ability to address hallucinations effectively. Moreover, neglecting the role of visual input will result in overfitting to language priors. In contrast, OViP (b) combines online preference learning with image-aware training in a unified framework, enabling real-time data construction grounded in model behavior.
}
\label{fig:intro}
\end{figure}

Building upon this negative generation strategy, we further introduce an online vision-language preference learning framework (OViP) on both response and image sides, to dynamically construct and learn from preference data during training. Similar to reinforcement learning paradigms, OViP samples LVLMs' outputs throughout the training process and creates both response-centric and image-centric preference pairs in real time. As illustrated in~\autoref{fig:intro}, these dual signals supervise the model to generate outputs more faithfully grounded in visual content. 
By continuously sampling and integrating new preference pairs based on emerging failure patterns, OViP enables adaptive learning that aligns with the evolving output distributions of the LVLMs. This ongoing adaptation mitigates the limitations of static datasets and reduces the reliance on extensive manual curation

We evaluate our framework on diverse multi-modal benchmarks, covering both hallucination-focused and general tasks. Based on our experiments, we find a notable trade-off between hallucination suppression and general capability or informativeness~(what we refer to as ``implicit hallucination''). To address this, we refine existing evaluation protocols and reveal that many prior methods tend to overestimate their improvements. Experimental results show that \ovip{} delivers significant advantages in both performance and efficiency. Furthermore, we investigate the role of online training and visual signals, as well as their interactions, in shaping training effectiveness.

%% file: sections/method.tex

\begin{figure}[t]
        \centering
        \includegraphics[width=\linewidth]{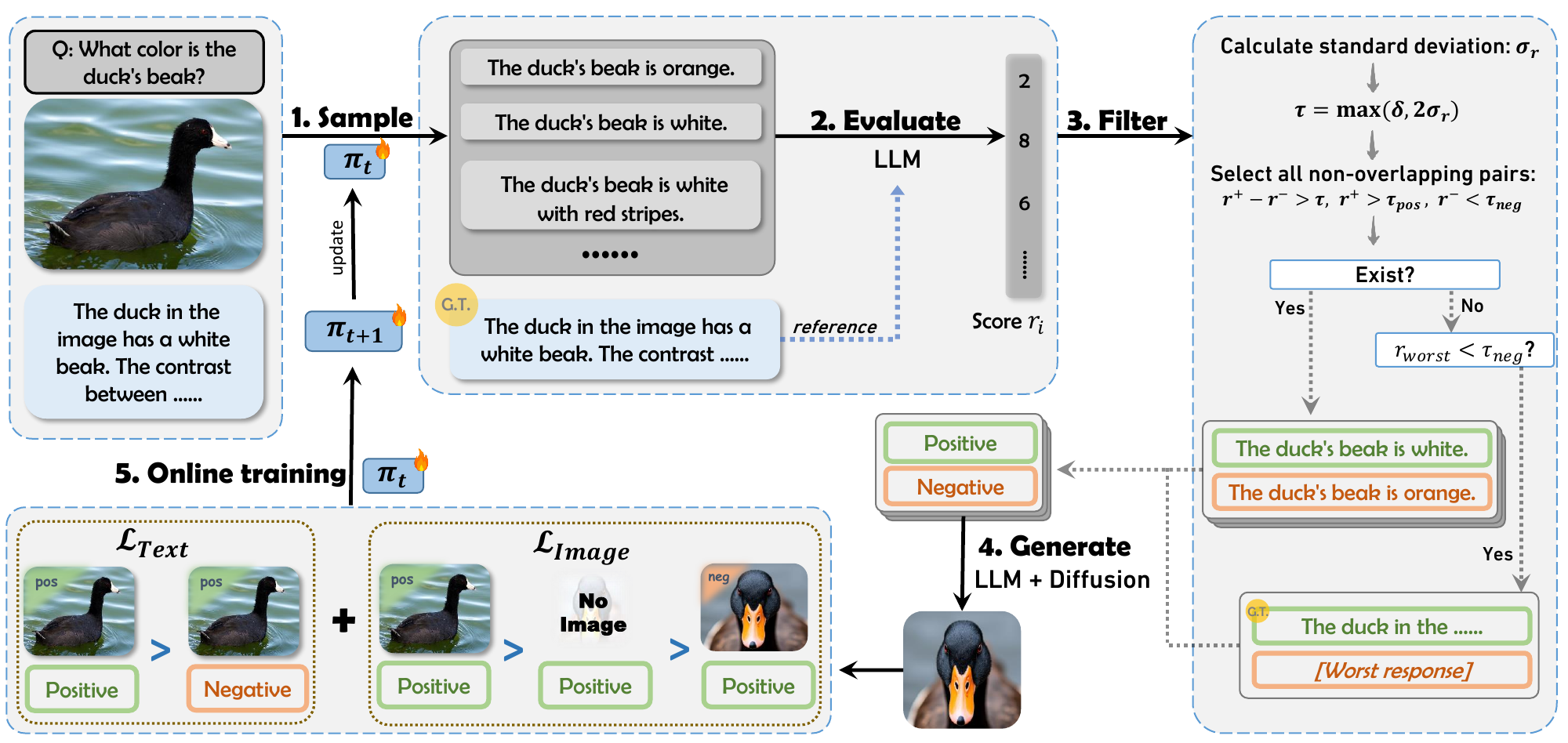}
        \caption{Overview of \ovip{}. Given an image and a query, we employ the current model~$\pi_t$ to generate multiple responses, which are then evaluated by an external LLM with reference to the ground truth. We filter and select response pairs and then generate corresponding negative images. The collected data are used to update $\pi_t$. The filtering strategy is detailed in Section~\ref{section:data_construction}.}
        \label{fig:main}
\end{figure}

In this section, we first provide an overview of the Online Vision-Language Preference Learning (OViP) framework (Section~\ref{section:overview}). We then elaborate the process of constructing the online preference pairs during training (Section~\ref{section:data_construction}) followed by how to learn from these preference data (Section~\ref{sec.loss-func}).

\subsection{Overview}
\label{section:overview}
As illustrated in Figure~\ref{fig:main}, our \ovip{} framework is designed to dynamically construct real-time preference pairs during training, by collecting in-distribution success and failure responses along with their corresponding original and synthesized images. These preference pairs are then integrated into the next training iteration for direct preference optimization on both image and response sides, providing a continuous feedback loop that refines the model’s visual grounding and improves its ability to distinguish high-quality outputs from suboptimal ones.

Specifically, given an input image $\mathcal{I}^+$, an instruction $\mathcal{Q}$, and a reference response $\mathcal{A}^*$, \ovip{} first samples multiple candidate responses using the target model $\pi$. These responses are then filtered and selected to form positive and negative pairs~$(\mathcal A^+,\mathcal A^-)$. Based on the semantic discrepancies between the response pairs, contrastive images~$\mathcal I^-$ are further synthesized to describing the negative responses. Finally, both image-level and response-level contrastive losses are applied to update the target model $\pi$. A detailed workflow of the OViP algorithm is provided in~\autoref{tab:app:algorithm}.

\subsection{In-Distribution Preference Data Construction}
\label{section:data_construction}
We adopt training-time inference to dynamically construct richer preference pairs reflecting in-distribution failures concurrently with the training process, expanding limited and static offline datasets. Specifically, our method involves three integral stages: (1) real-time generation of candidate outputs given visual inputs and instructions, (2) quality-aware sampling of informative preference pairs, and (3) inverse construction of input data conditioned directly on these sampled outputs. 
To ensure training stability, we implement dynamic sampling techniques and an experience buffer.

\paragraph{Real-time Generation of Output Data}
At each training step $s$, given a visual input $\mathcal I^+$ and its corresponding textual instruction $\mathcal Q$, our model $\pi_s$ generates  $k=16$ candidate responses $\mathcal A^i\ \left(i=1,2,\dots,k\right)$ through stochastic sampling. Each generated response is then individually evaluated by an LLM-based reward function~(denoted as $\mathrm G_\mathrm{r}$), which assigns a numerical reward score to each response, reflecting its alignment with the ground-truth answer $\mathcal A^*$. 
\begin{equation}
    \mathcal A^i\sim\pi_s\left(\cdot|\mathcal I^+,\mathcal Q\right);\quad r^i=\mathrm G_\mathrm{r}\left(\mathcal A^i,\mathcal A^*\right)
\end{equation}
\paragraph{Contrasting Response Pair Sampling}
To effectively learn from preference data, it is crucial to construct pairs with sufficient contrast in quality~\citep{dapo}. We dynamically construct preference pairs by selecting response pairs within each batch that display significant score disparities. Specifically, for each set of candidate responses $\{\mathcal A^i\}_{i=1}^k$ with corresponding rewards $\{r^i\}_{i=1}^k$, we compute the standard deviation $\sigma_r$ of the reward scores and select pairs $\left(\mathcal A^+, \mathcal A^-\right)$ that satisfy $|r^+ - r^-| > \max\left( \delta, 2\sigma_r \right)$
where $\delta$ is a fixed lower-bound margin. This criterion ensures that only response pairs exhibiting substantial contrast in reward scores are selected, effectively emphasizing informative differences between success and failure responses.
Additionally, we enforce quality constraints by requiring that the accepted positive responses meet a predefined quality criterion (i.e., $r^+ > \tau_{\text{pos}}$), while rejected negative responses fall below a specified threshold (i.e., $r^- < \tau_{\text{neg}}$). In cases where all candidate responses collectively perform poorly, we leverage offline ground-truth answers $\mathcal A^*$ as positive responses to guide the model learning effectively, a practice reminiscent of the mixed-policy approach in~\citet{onofflearning}.
\begin{equation}
\begin{aligned}
    \mathcal D_{\text{pair}} = \big\{ &(\mathcal Q, \mathcal I^+, \mathcal A^+, \mathcal A^-) \ \big|\ 
    \mathcal A^+, \mathcal A^- \in \{\mathcal A^i\}_{i=1}^k, \\
    &|r^+ - r^-| > \max(\delta, 2\sigma_r), r^+ > \tau_{\mathrm{pos}},\ r^- < \tau_{\mathrm{neg}} \big\}
\end{aligned}
\end{equation}


\paragraph{Inverse Negative Image Synthesis}
After obtaining response pairs $(\mathcal Q,\mathcal I^+,\mathcal A^+, \mathcal A^-) \in \mathcal D_{\text{pair}}$, we synthesize negative images corresponding to negative responses while taking input images as positive. 
Specifically, we utilize an external LLM~(denoted as $\mathrm{G}_{\text{diff}}$) to identify a set of semantic differences between the positive and negative responses, including entities, attributes, and spatial relations, and then generate a textual description $\mathcal T^-=\mathrm{G}_{\text{diff}}(\mathcal Q,\mathcal A^+, \mathcal A^-)$ that encapsulates the semantic content of the negative response $\mathcal A^-$. Subsequently, a diffusion-based image generation model~(denoted as $\mathrm{Diff}$) synthesizes a hard negative image as follows:
\begin{equation}
    \mathcal I^- = \mathrm{Diff}(\mathcal T^-)
\end{equation}
This inverse generation process, in which the image is conditioned on the textual output, ensures that the synthesized image captures hallucinated or incorrect content, providing more targeted supervision for hallucination mitigation. Moreover, as the generation is explicitly driven by response-level discrepancies, the resulting negative images exhibit higher semantic relevance and visual specificity.

\paragraph{Dynamic Inference and Experience Buffer}
To stabilize batch-wise training while retaining the flexibility of online sampling, we maintain an experience buffer $\mathcal{B}$ that stores dynamically constructed contrastive training samples. At each training step, the current model $\pi_s$  performs inference and response sampling, producing contrastive samples that are continuously added to $\mathcal{B}$. This sampling process persists until the accumulated samples reach the predefined batch size $N$. Once $\lvert\mathcal{B}\rvert \geq N$, a batch of $N$ samples is retrieved from $\mathcal{B}$ for loss computation and gradient updates. The remaining samples in the buffer are preserved for subsequent iterations, ensuring the training process to proceed smoothly even under variable sampling yields.

\subsection{Image- and Response-Side Preference Optimization}\label{sec.loss-func}

To effectively align both textual and visual modalities during training, we formulate a unified optimization framework that simultaneously considers response-level and image-level preference signals. 
The overall optimization objective consists of two complementary components. The first is the text DPO loss\citep{dpo}, which guides the model to learn response-level preferences conditioned on the input image and instruction:
\begin{equation}
    \mathcal{L}_{\mathrm{Text}}\left(\mathcal A^+,\mathcal A^-;\mathcal I^+,\mathcal Q\right) = -\log \sigma\left( \beta \cdot \left[ \log \frac{\pi_\theta(\mathcal A^+|\mathcal I^+,\mathcal Q)}{\pi_{\mathrm{ref}}(\mathcal A^+|\mathcal I^+,\mathcal Q)} - \log \frac{\pi_\theta(\mathcal A^-|\mathcal I^+,\mathcal Q)}{\pi_{\mathrm{ref}}(\mathcal A^-|\mathcal I^+,\mathcal Q)} \right] \right)
\label{eq:text-dpo}
\end{equation}

In addition to response-level alignment, we incorporate a contrastive objective focused on the visual input. By keeping the query and response fixed, the model is required to learn preferences solely from differences in the visual input. On top of this, to further ensure that the model's output maintains a reasonable and smooth probability distribution, we introduce the image-free term~$\pi_\theta(\mathcal A|\mathcal Q)$ and implement the image-side loss as in~\citet{svco}:
\begin{equation}
\begin{aligned}
\mathcal{L}_{\mathrm{Image}}(\mathcal I^+,\mathcal I^-; \mathcal Q,\mathcal A^+) = - \log \sigma\bigg( 
&\beta_1 \cdot \left[ \log \frac{ \pi_\theta(\mathcal A^+ | \mathcal I^+, \mathcal Q) }{ \pi_{\mathrm{ref}}(\mathcal A^+ | \mathcal I^+, \mathcal Q) } - 
\log \frac{ \pi_\theta(\mathcal A^+ | \mathcal Q) }{ \pi_{\mathrm{ref}}(\mathcal A^+ | \mathcal Q) } \right] \\
+ &\beta_2 \cdot \left[ \log \frac{ \pi_\theta(\mathcal A^+ | \mathcal Q) }{ \pi_{\mathrm{ref}}(\mathcal A^+ | \mathcal Q) } - 
\log \frac{ \pi_\theta(\mathcal A^+ | \mathcal I^-, \mathcal Q) }{ \pi_{\mathrm{ref}}(\mathcal A^+ | \mathcal I^-, \mathcal Q) } \right] \bigg)
\end{aligned}
\label{eq:image-dpo}
\end{equation}

The overall loss function is then defined as:
\begin{equation}
\begin{aligned}
\mathcal{L}_{\mathrm{OViP}}\left(\mathcal Q,\mathcal I^+,\mathcal I^-,\mathcal A^+,\mathcal A^-\right)=\mathcal{L}_{\mathrm{Text}}\left(\mathcal A^+,\mathcal A^-;\mathcal I,\mathcal Q\right)+\mathcal{L}_{\mathrm{Image}}\left(\mathcal I^+,\mathcal I^-; \mathcal Q,\mathcal A^+\right)
\end{aligned}
\label{eq:ovip-loss}
\end{equation}

%% file: sections/experiment.tex
\subsection{Experimental Setup}\label{Sec.exp.setup}
\paragraph{Implementation Details}
We conduct our experiments on LLaVA-1.5-7B-hf
and LLaVA-1.5-13B-hf~\citep{llava15}, with CLIP ViT-L-336px as the visual encoder and Vicuna-7b/13b as the backbone respectively. The training dataset, sourced from~\citet{opadpo}, consists of 8,730 samples and 4,013 distinct image–query combinations, including image description, question answering, and some yes/no questions. 
We use LoRA~\citep{lora} with a rank of 256 and alpha of 512. 
Other settings are listed in Appendix~\ref{app.sub.settings}

\paragraph{Baselines} We compare \ovip{} with SFT, DPO~\citep{dpo}, mDPO~\citep{mdpo} and GRPO~\citep{grpo}. As the original versions of SFT, DPO and mDPO are offline methods, we additionally implement iterative DPO and GRPO to facilitate a more comprehensive comparison. Furthermore, we evaluate several prior works with publicly available model weights, including HA-DPO~\citep{hadpo}, HALVA~\citep{halva}, RLAIF-V~\citep{rlaifv} and OPA-DPO~\citep{opadpo}. Among them, our OViP and OPA-DPO use the same original training data, which is a subset of the dataset used by RLAIF-V.

\subsection{Evaluation Metrics} \label{sub.evaluation_metrics}
We conduct evaluations on five \textbf{hallucination-related} and four \textbf{general capability} benchmarks to assess hallucination mitigation and overall capability degradation.

\paragraph{Hallucination-Related Evaluation.}
We evaluate hallucination in LVLM outputs using MMHal-Bench~(MMHal)~\citep{mmhal}, AMBER generative~(AMB$_\mathrm{gen}$)~\citep{amber}, Object HalBench~(ObjectHal)~\citep{objecthal}, Llava-Bench-in-the-Wild~(LV)~\citep{llavabenchinthewild}, and AMBER discriminative~(AMB$_\mathrm{dis}$)~\citep{amber}. Detailed descriptions of the datasets, evaluation procedures, and metrics are provided in Appendix~\ref{app.evaluation.benchmarks}

Prior work has primarily focused on assessing the precision of model outputs, i.e., whether the generated content contains explicit hallucinations. However, this perspective often overlooks the \emph{completeness} of the output: a model may omit relevant entities~(especially in image description tasks), leading to what we term \emph{implicit hallucinations}. \textit{We argue that both explicit and implicit hallucinations are critical for a faithful evaluation of model reliability.} 
Building on this perspective and the observation of failure cases where existing benchmarks can be hacked, we \textbf{refine the evaluation protocols and introduce an F1 score for AMB$_\mathrm{gen}$ and ObjectHal to better capture the extent of hallucination in generated responses}. Illustrative failure cases of prior evaluation strategies are presented in Appendix~\ref{app.evaluation.bad-case}.

To aggregate performance across five benchmarks, we introduce the \textbf{Hallucination Reduction Index}~(\textbf{HRI}) as a unified measure of overall improvement. HRI is computed by summing the normalized improvements from each benchmark to obtain the overall relative gain. The detailed calculation of HRI and the discussion of its fairness are provided in Appendix~\ref{app.evaluation.hri}.

\paragraph{General Capability Evaluation} To assess the trade-off between hallucination mitigation and general visual capability, we evaluate the trained models on general benchmarks, including RealworldQA~\citep{realworldqa}, TextVQA~\citep{textVQA}, CVBench~\citep{cvbench}, MMStar~\citep{mmstar}. 
We aggregate the results across these benchmarks and compute the \textbf{Accuracy Difference}, serving as a unified metric to quantify overall performance variation after training.

\subsection{Main Results}

\begin{table}[t]
\centering
\scriptsize
\caption{\textbf{Main Results for \ovip{} and other methods across different benchmarks.} The five \colorbox{gray!30}{shaded} metrics highlight the primary balanced and overall results for each benchmark. \colorbox{yellow!40}{HRI}~(Hallucination Reduction Index) is the average improvement across five benchmarks. Acc$_{\texttt{Dif}}$ is the total accuracy changes across TextVQA\citep{textVQA}, RealworldQA\citep{realworldqa}, MMStar\citep{mmstar} and CVBench\citep{cvbench}. \texttt{GPT4-V}($^\dagger$)'s results are cited from \citet{HSA-DPO},\citet{amber},\citet{leaderboard} for reference. $^\ddagger$ indicates the use of original evaluation strategy. 
$^*$ denotes methods with publicly released model weights trained on their own datasets, which we direct evaluate without re-training. $^\sharp$ signifies methods trained on datasets that are the same as or larger than ours. ``2-ep'' specifies results obtained after two epochs of training. We separate offline methods from non-offline methods for clearer comparison. }
\begin{tabular}{@{}clcccccccc|cc@{}}
\toprule
 &              & \multicolumn{3}{c}{AMB$_{\mathrm{gen}}$}       & MMHal & \multicolumn{2}{c}{ObjectHal} & LV           & AMB$_{\mathrm{dis}}$  &
 \multirow{2}{*}{\begin{tabular}{@{}r@{}}\cellcolor{yellow!40}\textbf{HRI}\end{tabular}}
 &General        \\ 
 &              & Chair$\downarrow$      & Cover$\uparrow$        & \cellcolor{gray!30}\textbf{F1}$\uparrow$           & \cellcolor{gray!30}\textbf{Score}$\uparrow$ & Chair$_r\downarrow$        & \cellcolor{gray!30}\textbf{F1}$\uparrow$          & \cellcolor{gray!30}\textbf{Score}$\uparrow$        & \cellcolor{gray!30}\textbf{F1}$\uparrow$     &    & Acc$_{\texttt{Dif}}$   \\ \midrule
 & \texttt{GPT4-V$^\dagger$} & 4.6 & 67.1 & 78.8\phantom{0} & \phantom{$^\ddagger$}3.49$^\ddagger$ & 13.6\phantom{0} & - & 95.3\phantom{0} & 87.4 & - & -\\
 \midrule
\multirow{14}{*}{\rotatebox{90}{\textbf{LLaVA-1.5-7B}}}
& Baseline     & 7.1 & 50.0 & 65.01 & 1.90  & 51.38           & 72.40       &57.20&85.5&-&-\\
&\textsc{HA-DPO}$^*$&5.6&49.4&64.86&1.95&37.15&73.81&57.30&85.4&\phantom{-}1.52&\textit{-11.59}\phantom{-}\\
&\textsc{HALVA}$^*$&5.7&52.9&\textbf{67.78}&2.12&43.40&\textbf{76.01}&58.60&86.5&\phantom{-}\textbf{9.08}&\textit{-7.36}\\
&\textsc{RLAIF-V}$^{*\sharp}$&3.1&49.8&65.79&2.54&\phantom{0}9.35&69.78&58.90&86.4&\phantom{-}1.37&\textit{-6.74}\\
&\textsc{OPA-DPO}$^{*\sharp}$&2.4&45.2&\textit{61.79}&\textbf{2.78}&\phantom{0}6.37&\textit{63.26}&\textbf{64.80}&86.7&\textit{-5.60}&\textit{-11.82}\phantom{-}\\
&SFT&3.5&50.6&66.39&2.52&20.60&70.30&\textit{52.20}&86.1&-1.47&\textit{-8.07}\\

&DPO&3.7&48.9&64.86&2.35&26.60&71.95&56.70&86.8&\phantom{-}1.65&-3.86\\
&mDPO&3.4&48.6&64.67&\textbf{2.55}&25.45&\textbf{73.92}&55.80&86.1&\phantom{-}2.99&-3.05\\
\cline{2-12}\rule{0pt}{2.5ex}&DPO$_{\mathrm{iterative}}$&3.9&48.7&64.64&2.32&27.11&72.33&56.40&86.5&\phantom{-}1.31&-2.98\\
&GRPO$_{2\mathrm{ep}}$&4.8&51.2&66.59&2.45&34.98&73.83&58.70&86.8&\phantom{-}6.75&-3.83\\
&\textbf{\ovip}&4.0&51.1&\textbf{66.70}&2.52&33.22&73.50&\textbf{63.10}&\textbf{87.3}&\phantom{-}\textbf{9.58}&\textbf{\phantom{-}+0.88\phantom{+}}\\
&\textbf{\ovip}$_{2\mathrm{ep}}$&4.0&51.6&\textbf{67.12}&\textbf{2.65}&29.54&\textbf{74.18}&\textbf{60.90}&\textbf{87.4}&\textbf{10.00}&\textbf{-1.01}\\\midrule
\multirow{9}{*}{\rotatebox{90}{\textbf{LLaVA-1.5-13B}}}
&Baseline&6.5&51.0&65.99&2.24&46.18&76.73&62.60&89.1&-&-\\
&\textsc{HALVA}$^*$&6.0&52.2&67.12&2.45&35.07&\textbf{77.75}&61.70&90.0&\phantom{-}4.22&\textit{-5.45}\\
&\textsc{OPA-DPO}$^{*\sharp}$&2.8&47.8&64.08&\textbf{2.88}&\phantom{0}5.88&\textit{64.46}&64.70&89.3&\textit{-7.05}&\textit{-15.25}\phantom{-}\\
&SFT&4.5&50.0&65.64&2.38&31.21&75.81&64.00&89.9&\phantom{-}1.79&-1.24\\
&DPO&3.6&50.6&66.37&2.53&25.00&75.00&65.30&89.6&\phantom{-}2.42&\phantom{-}+0.12\phantom{+}\\
&mDPO&3.9&50.1&65.86&2.51&21.79&75.35&64.50&89.5&\phantom{-}1.78&-1.12\\
\cline{2-12}\rule{0pt}{2.5ex}&GRPO$_{2\mathrm{ep}}$&3.8&52.4&67.84&2.38&23.76&75.55&\textbf{66.70}&\textbf{90.4}&\phantom{-}4.96&-1.48\\
&\textbf{\ovip}&4.4&53.1&\textbf{68.28}&\textbf{2.58}&36.30&76.52&64.60&89.7&\phantom{-}\textbf{5.25}&\phantom{-}\textbf{+0.85}\phantom{+}\\
&\textbf{\ovip}$_{2\mathrm{ep}}$&3.6&53.7&\textbf{68.98}&2.57&28.62&\textbf{76.75}&\textbf{67.90}&\textbf{90.2}&\phantom{-}\textbf{8.02}&\phantom{-}\textbf{+2.02}\phantom{+}\\
\bottomrule
\end{tabular}\label{tab:main}
\end{table}

\autoref{tab:main} presents results for \ovip{} and other methods across multiple benchmarks on various LVLM backbones. \textbf{\ovip{} consistently achieves significant improvements across most primary metrics while effectively preserving the model's general visual capabilities}~(achieving +0.88 with one epoch for General Acc$_\mathrm{Dif}$ and a slight drop of -1.01 for 2 epochs), whereas most other methods that exhibit varying degrees of degradation in general benchmarks. Moreover, \ovip{} further improves with an additional training epoch. Notably, even with one epoch, \ovip{} surpasses HALVA and 2-epoch GRPO, both of which utilize twice as much training data, but still yield lower HRI and suffer from general ability degradation.



A critical phenomenon often overlooked in previous work~\citep{v-dpo,opadpo,chip,HSA-DPO,mdpo,rlhfv,rlaifv} is that \textbf{offline methods generally impair models’ general capability while also introducing implicit hallucinations}~(as discussed in \autoref{sub.evaluation_metrics}). This issue is particularly evident in OPADPO, where Chair score on AMB$_\mathrm{gen}$ drops to 2.4, and Cover metric decreases from the initial 50.0 to 45.2, far below other methods. An illustrative example of such omission is in \autoref{fig:app.bad_cases.amber.omit} in Appendix. Moreover, excessive training further exacerbates this problem: as shown in \autoref{tab:main}, several DPO-like methods ~(HA-DPO, HALVA, RLAIF-V, OPA-DPO) trained for more than two epochs suffer from much larger declines in general capability compared to DPO and mDPO trained for only one epoch. At the same time, except for HALVA, their HRI scores are also lower than those of DPO and mDPO, which mainly influenced by the low F1 scores on AMB$_\mathrm{gen}$ and ObjectHal. \textit{With these possible signs of overfitting, we suggest that some improvements reported in prior work may be overestimated}.

\subsection{Ablation Study}
\paragraph{The Impact of Loss functions.}



We evaluated various combinations of loss functions for online preference learning in hallucination mitigation to derive the final formulation in \autoref{eq:ovip-loss}. Our ablation study examines the effectiveness of different training objectives, including text-side ($\mathcal{L}_{\mathrm{Text}}$), image-side and auxiliary losses. Specifically for image-side losses, we examine our image loss~$\mathcal L_{Image}$ alongside two variants $\mathcal L_{Image}^{base}$ and $\mathcal L_{Image\mathrm{-}Sym}$. For auxiliary loss, we compare the anchor loss proposed by~\citet{mdpo} and the bidirectional anchor loss, which enforce the probability of positive response to increase and the negative one to decrease. Detailed formulations are provided in Appendix~\ref{app.sub.loss-function}.

\begin{table}[!ht]
\centering
\begin{minipage}{0.45\linewidth}
    \caption{Results of different loss functions. $\mathcal L_\mathrm{OViP}=\mathcal L_\mathrm{Text}+\mathcal L_\mathrm{Image}$. 
    }
\footnotesize
\begin{tabular}{@{}lccc@{}}
\toprule
\multirow{2}{*}{Loss Functions}&\multicolumn{2}{c}{\textbf{HRI}}&\\
\cline{2-4}\rule{0pt}{2.5ex}
& \textit{From Scratch} & \textit{Iterative}&\\
\midrule
$\mathcal L_\mathrm{OViP}$&\phantom{-}\textbf{4.32}&\textbf{7.94}&\\
\ \ \ \ $-\ \mathcal L_\mathrm{Text}$&\phantom{-}4.23&7.71&\\
\ \ \ \ $-\ \mathcal L_\mathrm{Image}$&-2.29&4.56\\
$\mathcal L_\mathrm{Text}+\mathcal{L}_\mathrm{Image}^{\ base}$&\phantom{-}4.08&7.50\\
$\mathcal L_\mathrm{Image-Sym}$&-0.32&6.57\\
\bottomrule
\end{tabular}\label{tab:abl.loss}
\end{minipage}
\hfill
\begin{minipage}{0.5\linewidth}
\caption{Results of offline and online training strategy with DPO and \ovip{}. Cover measures the informativeness of the model from AMB$_\mathrm{gen}$. Cover score of the original model is 50.}
\footnotesize
\begin{tabular}{@{}llccc@{}}
\toprule
\multicolumn{2}{c}{Method} & \textbf{Cover} & \textbf{HRI} & \textbf{General Acc$_{\mathrm{Dif}}$}\\
\midrule
\multirow{2}{*}{OViP} & online & \textbf{51.1} & \phantom{-}\textbf{9.36} & \phantom{-}\textbf{0.88} \\
 & offline & 49.9 & \phantom{-}4.32 & \phantom{-}0.08 \\
\midrule
\multirow{2}{*}{DPO} & online & \textbf{50.0} & \phantom{-}\textbf{1.71} & -2.57 \\
 & offline & 48.3 & -2.29 & -\textbf{1.38} \\
\bottomrule
\end{tabular}\label{tab:abl.online}
\end{minipage}
\end{table}

We conduct experiments under two training regimes: (1) training from scratch, and (2) iterative training initialized with a DPO-pretrained model using the existing dataset, to ablate these losses on top of different initialized models with varying capabilities. We observe that models trained with different losses do not suffer from a notable drop in general ability (\textbf{General Acc}$_\mathrm{Dif} > -1.5$). Therefore, in \autoref{tab:abl.loss} we only report the HRI results, which show that the full \ovip{} loss consistently outperforms all variants under both training regimes. Moreover, the form of the image loss greatly affects the results, with the loss in \autoref{eq:image-dpo} achieving the best performance.

\begin{figure}[htbp]
    \centering
    \begin{minipage}[b]{0.48\textwidth}
        \centering
        \includegraphics[width=\linewidth]{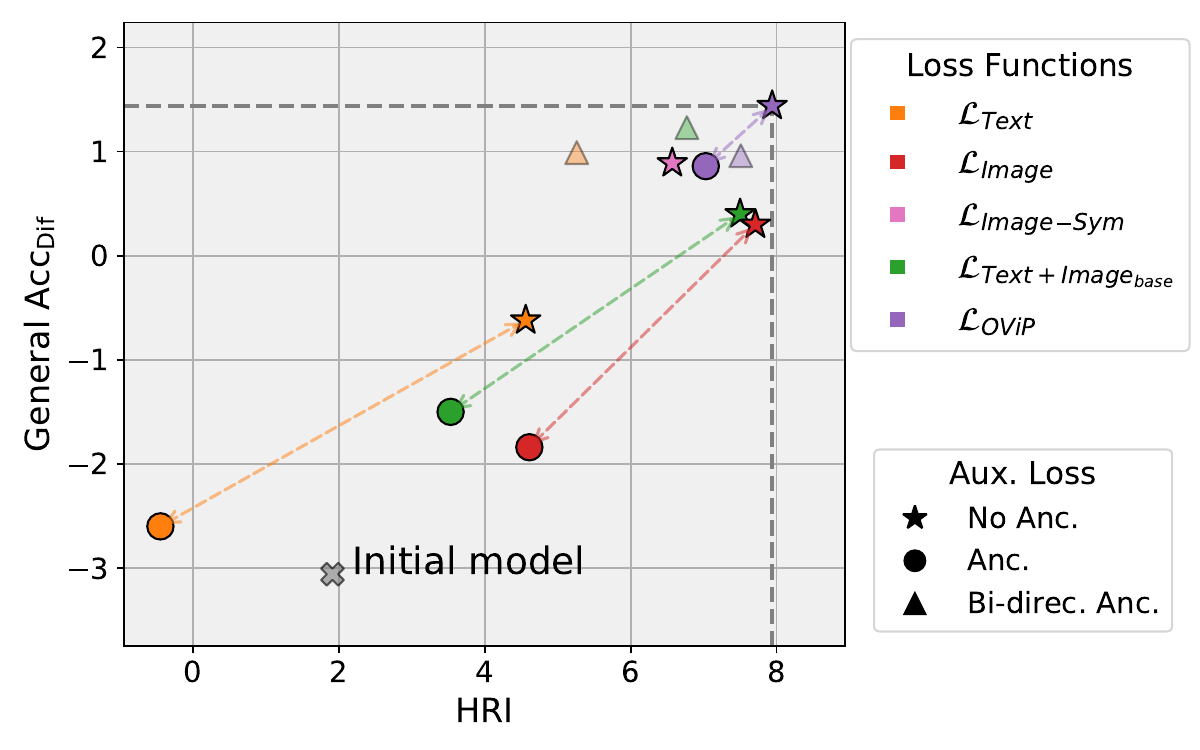}
        \caption{Effect of applying auxiliary loss to different loss functions.}
        \label{fig:abl.loss}
    \end{minipage}
    \hfill
    \begin{minipage}[b]{0.48\textwidth}
        Under the same iterative training regime, we further analyze the effect of auxiliary losses based on the DPO-initialized model and its sampled responses, as illustrated in~\autoref{fig:abl.loss}. Contrary to the findings in mDPO~\citep{mdpo}, we find that \textbf{incorporating anchor loss consistently reduces general capability and increases hallucinations across all loss combinations}. Moreover, while applying bi-directional anchor loss slightly improves general capabilities, it does not necessarily enhance hallucination mitigation. Therefore, \ovip{} loss without anchor loss is the most effective training objective for both reducing hallucination and maintaining general ability.
    \end{minipage}
\end{figure}


\paragraph{Online v.s. Offline.}
\autoref{tab:abl.online} demonstrates that online training consistently outperforms its offline counterpart in HRI by at least 4 points within just one epoch~(and continues to improve with further training, while offline training suffers from overfitting). 
Another notable observation is that \textit{online training also improves the informativeness of model outputs}. Even when trained with DPO, the Cover score remains 50. In contrast, previous studies~\citep{rlaifv,opadpo,chip} using the similar dataset typically exhibit a drop in this aspect.
Additionally, the improvement for online training over offline training is almost across every individual benchmark and each corresponding metric, online training yields more stable and superior performance. Detailed results are provided in Appendix~\ref{app.sub.OOresult}.

 

%
%

%% file: sections/further_study.tex
\subsection{Training Efficiency}\label{sub.training_efficiency}

\begin{figure}[ht]
    \begin{minipage}[b]{0.50\textwidth}
        \centering
        \includegraphics[width=\linewidth]{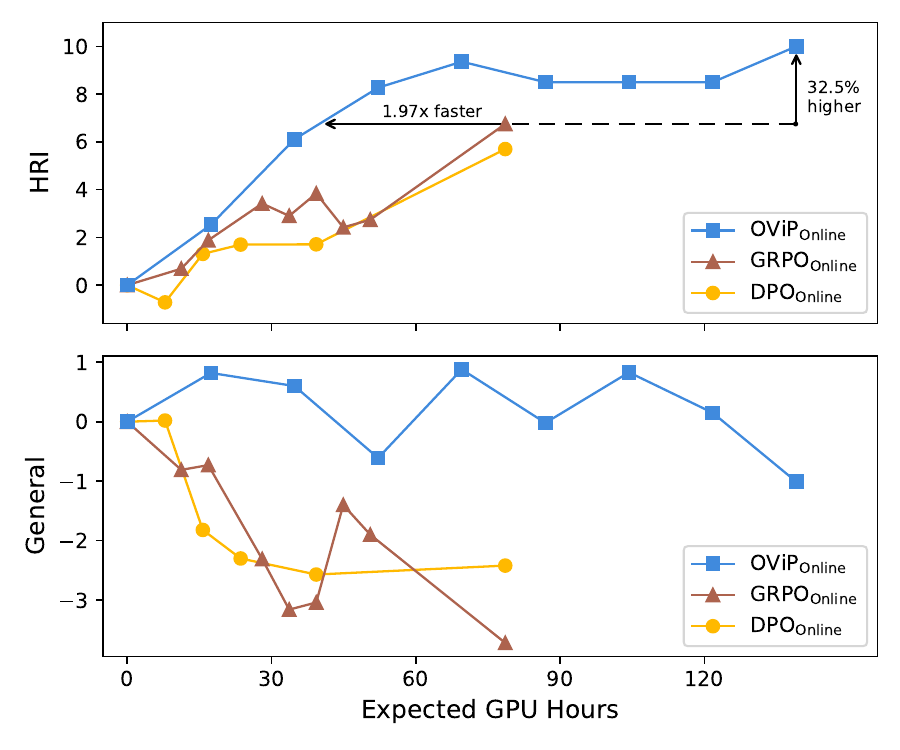}
        \caption{\textbf{Performance comparison among online training methods up to 2 epochs.} The X-axis shows the expected training time multiplied by the number of GPUs used. OViP outperforms GRPO with 1.97$\times$ higher training efficiency.}
        \label{fig:further.online_time}
    \end{minipage}
    \hfill
    \begin{minipage}[b]{0.45\textwidth}
        \centering
        \includegraphics[width=\linewidth]{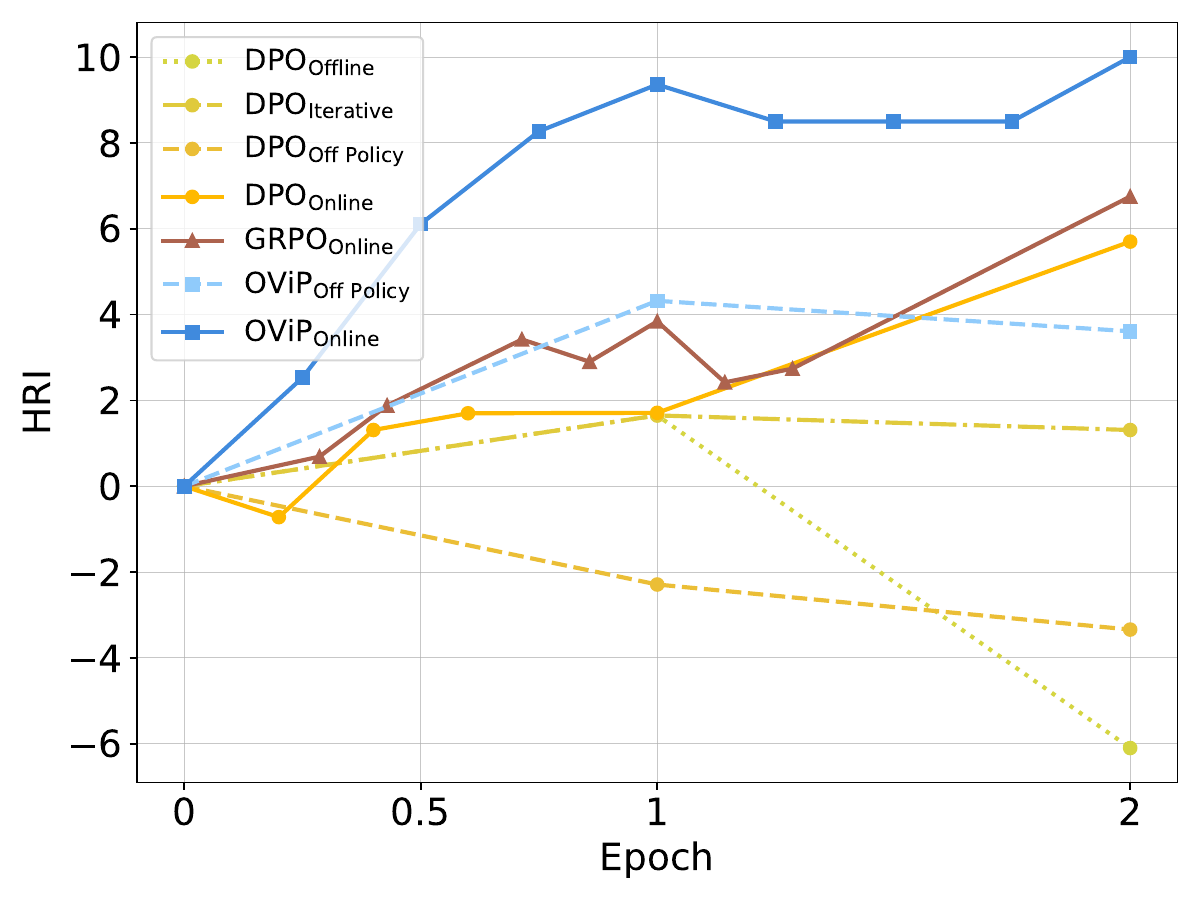}
        \caption{\textbf{Results under different training strategies}. Offline denotes training with the existing dataset; Off Policy refers to training with sampled data; and Iterative indicates that the dataset for the second epoch is generated by sampling from the offline-trained model after the first epoch. }
\label{fig:further.online}
    \end{minipage}
\end{figure}

Although \ovip{} requires constructing negative images, which needs additional GPU resources for deploying diffusion models and incurs extra time overhead, we show that \ovip{} still has clear advantages in overall training efficiency. In \autoref{fig:further.online_time}, we compare different online methods by plotting their HRI and general capability against expected GPU hours.~(A detailed analysis of training cost and efficiency is provided in Appendix~\ref{app.eff}) \textit{The results show that despite slower per-iteration speed, \ovip{} achieves approximately 1.97$\times$ higher training efficiency than GRPO}. \ovip{} requires only about half the computation of GRPO to achieve comparable performance, while Online DPO performs slightly worse than GRPO. As for offline approaches, although their data construction and training require a similar amount of computation, their performance consistently falls short of their online counterparts; hence, our efficiency comparison focuses on online methods.

\subsection{Training Dynamics}
\autoref{fig:further.online} illustrates how HRI evolves during training under different strategies, which allows us to investigate the dynamics of hallucination throughout training.

\paragraph{Need for Visual and Online Signals}
For hallucination mitigation in LVLMs, \textit{adding visual supervision signals proves crucial}: offline \ovip{} surpasses GRPO and all DPO variants with one epoch. Building on this, \textit{online methods offer further advantages, which not only make each optimization step more effective in reducing hallucinations, but also exhibit better scalability}, with overfitting arising significantly later compared to non-online approaches, whose performance starts to drop after training for one epoch. We conjecture that this superiority stems from the model-specific nature of hallucinations, which requires supervision to precisely target the current model’s errors.

\paragraph{Early Training Stagnation}
Both Online DPO and Off-Policy DPO exhibit an initial drop in performance, while GRPO and OViP show relatively slow improvement during the early stages of training. We attribute this phenomenon to the model’s initially skewed output distribution. Early training primarily increases the diversity of model outputs, which does not immediately translate into performance gains but expands the search space for subsequent learning. A detailed discussion is provided in Appendix~\ref{app:sub.training_dynamics}.

\begin{figure}
    \centering
    \includegraphics[width=0.7\linewidth]{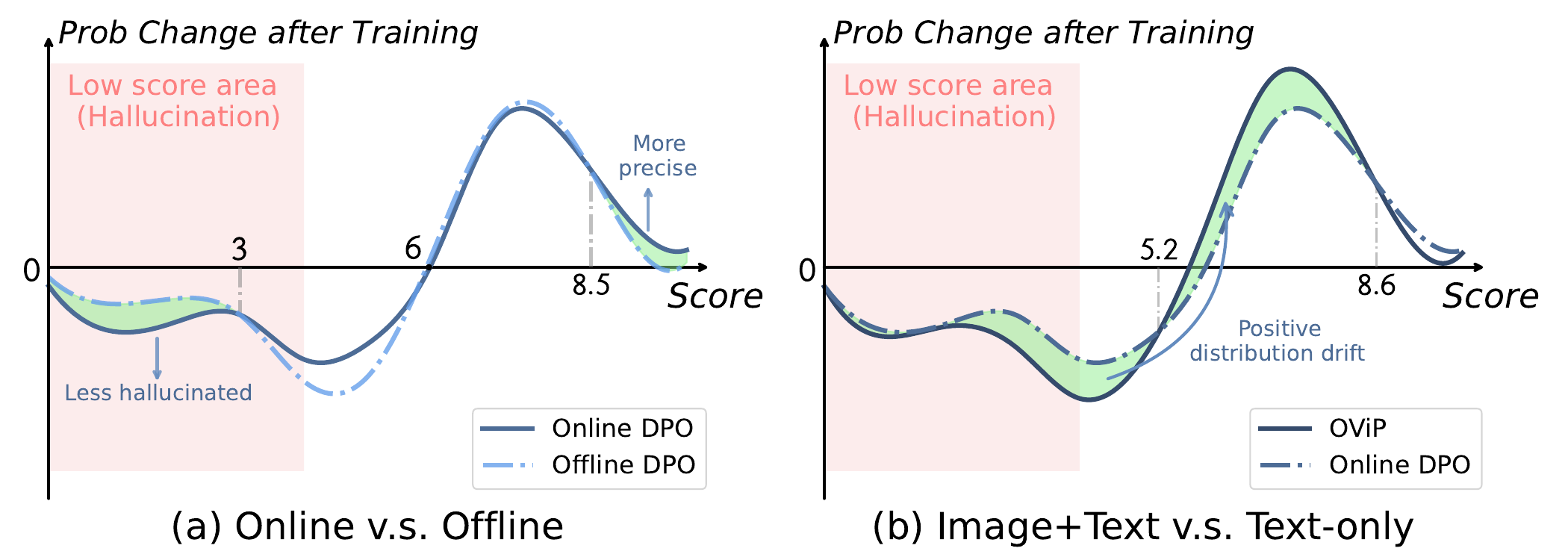}
    \caption{Change in probability mass for the responses with corresponding score after training. ``Low score'' refers to scores less than 4. We smooth the discrete probability changes over 11 score bins~(0-10) into a continuous curve. ``Change > 0'' represents the probability increases after training.}
    \label{fig:prob_change}
\end{figure}
\subsection{What Do We Actually Optimize during Training?}
We investigate how training affects the distribution of response quality, as shown in \autoref{fig:prob_change}. The results indicate that performance gains come from both reducing hallucinated outputs and increasing high-quality ones. After training, responses with scores below 6 become less likely, while those above 6 occur more frequently. 

\textit{Online training yields additional benefits over offline DPO. It further lowers the probability of severely hallucinated responses} (score < 3) \textit{and raises the probability of very high-quality ones} (score > 9), whereas offline DPO hardly changes the distribution at these extremes. \textit{Adding visual supervision signals shifts the overall distribution in a positive direction}, though it makes little extra difference to the probabilities at the lowest and highest scores. These observations suggest that the benefits of online training and those of visual supervision are orthogonal, thereby explaining OViP's stronger improvements achieved.

%% file: sections/related_work.tex

\subsection{LVLM Hallucination}


Works of synthetic data construction for mitigating hallucination in LVLMs can be broadly categorized into image-related synthesis and text-only synthesis. On the image side, several approaches leverage entity extraction and masking to perform targeted image editing, generating visually similar but semantically distinct counterfactuals~\citep{v-dpo,AdaViP2504}. In contrast, HallusionBench~\citep{hallusionbench} adopts a manual approach, carefully crafting counterfactual images to probe specific failure modes. Other works take a generative perspective: SynthVLM~\citep{synthvlm} and SynthEmbedding~\citep{synthembedding} utilize off-the-shelf models to synthesize new images or directly generate image embeddings for hallucination-aware training.
Meanwhile, text-side data augmentation can also be used in LVLM training. VoCoT~\citep{vocot} introduces new prompting patterns and response types to generate hallucination-prone QA data at scale. Other works such as~\citet{povid}, \citet{halva}, \citet{BDHS} introduce noise through perturbation, masking, or controlled corruption to simulate erroneous responses. More recent approaches~\citep{HSA-DPO,rlhfv} aim to detect and correct hallucinated content at varying levels of granularity, from token-level edits to full-sequence rewrites.

These efforts significantly improve the diversity and coverage of supervision signals available for training hallucination-robust VLMs.


\subsection{Allocating More Computation on Training Sample Construction}
Recent research has increasingly adopted the paradigm of allocating additional computation during training to get better training samples. Several studies utilize reinforcement learning with human or AI-generated feedback to guide VLM outputs. RLHF-V~\citep{rlhfv} leverages fine-grained human annotations to correct hallucinated content, while RLAIF-V~\citep{rlaifv} replaces human labels with feedback from ensembles of open-source models, significantly reducing annotation overhead. Similarly, OPA-DPO~\citep{opadpo} employs an on-policy editing step prior to DPO, aligning training samples closely with model predictions to enhance data efficiency. CLIP-based methods dynamically filter self-generated samples for high-quality training pairs \citep{CLIP-DPO,zhou2024calibratedselfrewardingvisionlanguage}. Other methods integrate auxiliary reward models or evaluators during training, providing continuous and adaptive feedback loops~\citep{mmhal,ViGoR}. Additionally, recent approaches incorporate reasoning or editing mechanisms directly into training, using iterative self-feedback or generative data augmentation techniques to dynamically refine model outputs~\citep{hadpo,ESREAL}. These strategies improve model alignment and factuality by enriching the quality and relevance of supervision signals during training.

%% file: sections/conclusion.tex
In this work, we propose the Online Vision-language Preference Learning (\ovip{}) framework to efficiently address the hallucination problem in LVLMs. By integrating online preference learning with image-aware training, \ovip{} enables real-time construction of high-quality contrastive data during training. Furthermore, to better assess the trade-offs between hallucination reduction and overall performance, we refine and extend existing evaluation protocols. Experimental results demonstrate that \ovip{} significantly outperforms prior offline/online training approaches, achieving substantial hallucination reduction while preserving general vision-language capabilities, which many existing offline methods fail to preserve. Our investigation into training dynamics also sheds light on the underlying mechanisms behind OViP’s effectiveness.

%% file: appendix/Ethic_statement.tex
This work focuses on improving the factual reliability of vision-language models by reducing hallucination. While it does not directly engage with societal applications, it contributes to the broader goal of building more trustworthy and robust AI systems. Although the method itself does not pose obvious risks, we note that even improved generation quality does not eliminate the possibility of misuse, such as producing misleading content. Responsible deployment and proper safeguards remain necessary when integrating such models into real-world applications.

%% file: appendix/evaluation.tex
\subsection{Hallucination Reduction Index}\label{app.evaluation.hri}
\subsubsection{Metric Design}
HRI represents an aggregate improvement metric across five different benchmarks. Simply summing the raw scores from each benchmark would not be a reasonable or rigorous approach, as the metrics are not directly comparable. Therefore, we calculate the improvement ratio for each benchmark based on its potential improvement range, effectively converting the raw metric gains into an additive proportion of improvement. Furthermore, we employ a conservative aggregation method to avoid overestimating the effectiveness of our approach.

Let $a_i,i\in\left\{1,2,3,4,5\right\}$ denotes $\mathrm{F1_\mathrm{AMB-gen},Score_{MMHal},F1_\mathrm{ObjectHal},LV_\mathrm{score},F1_{AMB-dis}}$ respectively, namely the results on each benchmark, superscript ``$^\mathrm{base}$'' represents performances of the baseline model and ``$^\mathrm{ref}$'' represents the set reference performances. Then HRI is calculated as:
\begin{equation}
    \mathbf{HRI}=2\times\sum_{i=1}^5\frac{a_i-a^{\mathrm{base}}_i}{a^{\mathrm{ref}}_i-a^{\mathrm{base}}_i}
\end{equation}

\subsubsection{Main Results}
For 7B model, we set the reference performances as \textbf{OViP}$_\mathrm{2ep}$, so it comes:
\begin{equation*}
    \mathbf{HRI}=2\times\left(\frac{a_1-65.01}{67.12-65.01}+\frac{a_2-1.90}{2.65-1.90}+\frac{a_3-72.40}{74.18-72.40}+\frac{a_4-57.20}{60.90-57.20}+\frac{a_5-85.5}{87.4-85.5}\right)
\end{equation*}

For 13B model, we also use \textbf{OViP}$_\mathrm{2ep}$ as the reference performances except for the ObjectHal benchmark which almost all methods fail to improve. We set the reference performance of ObjectHal to 79.0.
\begin{equation*}
    \mathbf{HRI}=2\times\left(\frac{a_1-65.99}{68.98-65.99}+\frac{a_2-2.24}{2.57-2.24}+\frac{a_3-76.73}{79.00-76.73}+\frac{a_4-62.60}{67.90-62.60}+\frac{a_5-89.1}{90.2-89.1}\right)
\end{equation*}

\subsubsection{Ablation Study: Loss Functions}
There is no method surpassing other methods significantly, so we consider the best performance on the benchmark as its reference peerformance.
\begin{equation*}
    \mathbf{HRI}=2\times\left(\frac{a_1-65.01}{68.57-65.01}+\frac{a_2-1.90}{2.70-1.90}+\frac{a_3-72.40}{74.14-72.40}+\frac{a_4-57.20}{64.10-57.20}+\frac{a_5-85.5}{87.20-85.5}\right)
\end{equation*}

\subsubsection{Ablation Study: Online and Offline}
Same as Main Results.
\begin{equation*}
    \mathbf{HRI}=2\times\left(\frac{a_1-65.01}{67.12-65.01}+\frac{a_2-1.90}{2.65-1.90}+\frac{a_3-72.40}{74.18-72.40}+\frac{a_4-57.20}{60.90-57.20}+\frac{a_5-85.5}{87.4-85.5}\right)
\end{equation*}

\subsubsection{Further Study}
Same as Main Results.
\begin{equation*}
    \mathbf{HRI}=2\times\left(\frac{a_1-65.01}{67.12-65.01}+\frac{a_2-1.90}{2.65-1.90}+\frac{a_3-72.40}{74.18-72.40}+\frac{a_4-57.20}{60.90-57.20}+\frac{a_5-85.5}{87.4-85.5}\right)
\end{equation*}
\subsubsection{Fairness}
When aggregating different metrics through weighted averaging, it is necessary to account for the relative importance of each metric. Here, we define the potential improvement of a metric by considering its maximum observed gain in comparable experiments, and assign its weight as the inverse of this gain to normalize across metrics. For example, if metric A shows a maximum improvement of 2 points while metric B improves by 4 points, we assume that an equally strong model would, on average, achieve only half as much gain on A as on B. Consequently, each point of improvement on A should be considered twice as important as a point on B. Compared with simple averaging, this weighting scheme better reflects the relative significance of different metrics and is less prone to being gamed.

\subsection{Benchmarks}\label{app.evaluation.benchmarks}
\begin{itemize}[leftmargin=2em] 
\item MMHal-Bench (MMHal)~\citep{mmhal} is a model-evaluated question-answering benchmark covering 8 categories and 12 topics. While the original evaluation strategy uses \texttt{GPT-4} to judge model responses, a text-only model will introduce considerable judging-time hallucinations and errors, so \texttt{gpt-4o-2024-05-13} is better for evaluation.~\citep{BDHS}. 
\item AMBER generative (AMB$_\mathrm{gen}$)~\citep{amber} is a judging-model-free benchmark for the image description task, comprising 1,004 samples. \textbf{Chair} measures the object-level hallucination rate as the average precision of objects mentioned in the model's descriptions, while \textbf{Cover} indicates the recall of objects. \textit{We observe a noticeable trade-off between these two metrics across various methods, where improvements in one often come at the expense of the other. To provide a more balanced and overall assessment, we introduce a new \textbf{F1} score calculated as the harmonic mean of Chair and Cover.} 
\item Object HalBench (ObjectHal)~\citep{objecthal} evaluates object-level completeness and hallucination rates. The generation prompts are augmented from~\citet{rlhfv}. \textbf{Chair$_r$} denotes the response-level hallucination rate. We also introduce an object-level \textbf{F1} metric to comprehensively measures the balance between hallucination and object coverage. Objects extraction is performed using \texttt{gpt-4o-2024-05-13}. 
\item Llava-Bench-in-the-Wild (LV)~\citep{llavabenchinthewild} evaluates models' visual abilities, using 60 open-ended questions grounded in 24 diverse images from real-world and abstract scenarios. The evaluation is conducted using \texttt{gpt-4o-2024-05-13}. 
\item AMBER discriminative (AMBER$_\mathrm{dis}$)~\citep{amber} includes 14,216 `Yes/No'' questions regarding objects in image. We use the \textbf{F1} score as its metrics. \end{itemize}

\subsection{Bad Cases}\label{app.evaluation.bad-case}
\subsubsection{MMHal}

\begin{figure}
    \centering
    \includegraphics[width=0.92\linewidth]{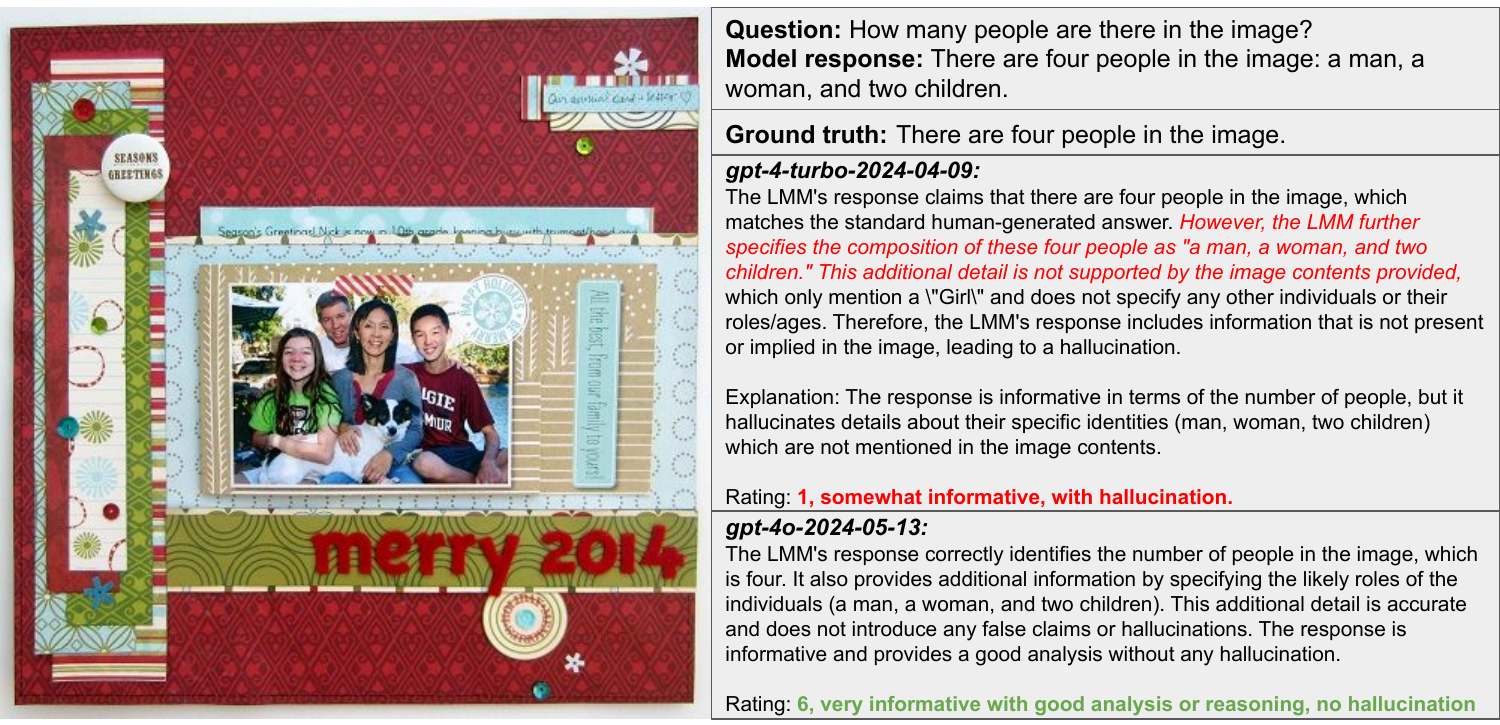}
    \caption{Text-only LLM can not correctly judge the response.}
    \label{fig:app.bad_cases.mmhal}
\end{figure}

Shown in \autoref{fig:app.bad_cases.mmhal}, the original evaluation protocol utilizes the text-only \texttt{gpt-4-turbo-2024-04-09} to evaluate the model response, which has no access to the input image and can only infer from the given image contents and ground truth, so it will lead to many incorrect judgments. We replace it with \texttt{gpt-4o-2024-05-13}, which yields more accurate assessments.
\subsubsection{AMBER-generative \& ObjectHal}
AMBER uses an automatic method for detecting the hallucination entity, which primarily relies on the pre-defined hallucination words. ObjectHal introduces LLM to extract the mentioned entities, its metrics are basically the same with AMBER.
\begin{figure}
    \centering
    \includegraphics[width=0.92\linewidth]{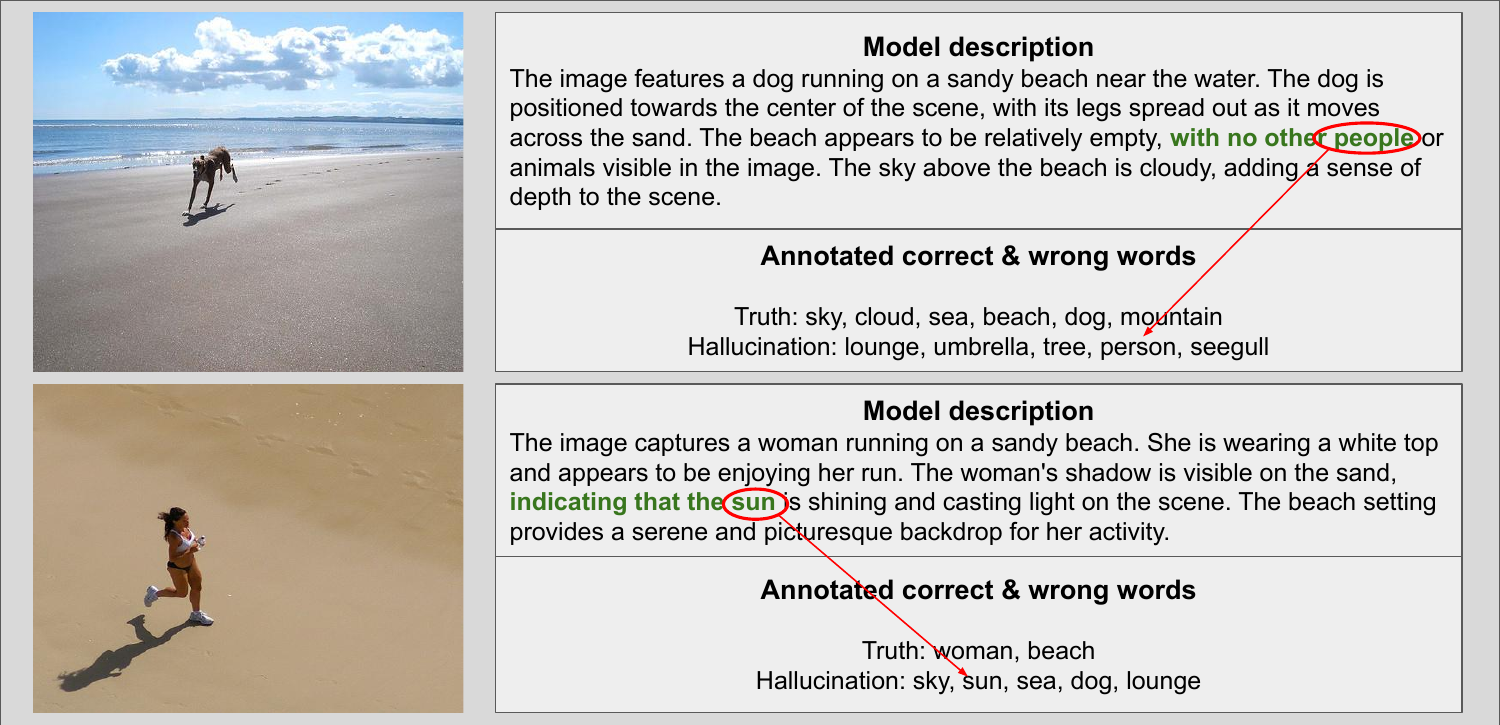}
    \caption{Rule-based extraction will lead to misjudgments to some extent.}
    \label{fig:app.bad_cases.amber.hal}
\end{figure}
\begin{figure}
    \centering
    \includegraphics[width=0.92\linewidth]{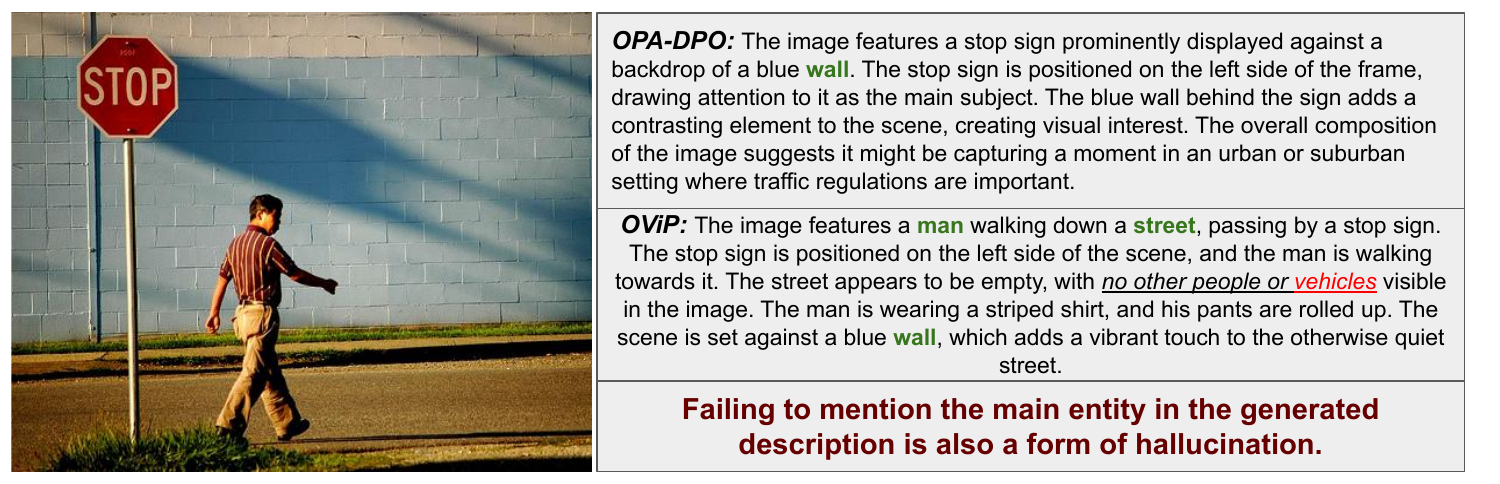}
    \caption{OPA-DPO fails to mention the man, a deficiency that is captured by Cover score but often overlooked in previous evaluations. ``\textcolor{red}{vehicles}'' is incorrectly identified as a hallucination word.}
    \label{fig:app.bad_cases.amber.omit}
\end{figure}
\autoref{fig:app.bad_cases.amber.hal} illustrates several cases of misjudgment in AMBER. Since the score is determined solely by the presence of specific predefined words rather than the actual semantic correctness, the hallucination rate (Chair score) is often overestimated. \textbf{Moreover, this issue becomes more pronounced as the diversity and informativeness of model responses increases.}

Many methods achieve great improvements in the Chair score (entity-wise hallucination rate), but often at the cost of a significant decrease in the cover rate (completeness and informativeness). \autoref{fig:app.bad_cases.amber.omit} provides an example of this information deficit phenomenon, which should also be considered in the evaluation of model performance.




%% file: appendix/ablation.tex
\subsection{Loss Functions}\label{app.sub.loss-function}

Base image loss~$\mathcal{L}_{\mathrm{Image}}^{base}$ is similar to DPO loss which replace the response pair with the image pair:
\begin{equation*}
    \mathcal{L}_{\mathrm{Image}}^{base}\left(\mathcal I^+,\mathcal I^-;\mathcal Q,\mathcal A^+\right) = \log \sigma\left( \beta \cdot \left[ \log \frac{\pi_\theta(\mathcal A^+|\mathcal I^+,\mathcal Q)}{\pi_{\mathrm{ref}}(\mathcal A^+|\mathcal I^+,\mathcal Q)} - \log \frac{\pi_\theta(\mathcal A^+|\mathcal I^-,\mathcal Q)}{\pi_{\mathrm{ref}}(\mathcal A^+|\mathcal I^-,\mathcal Q)} \right] \right)
\end{equation*}

Symmetrical image loss~$\mathcal{L}_{\mathrm{Image-Sym}}$ considers the negative image and the negative response a correct pair, then calculate Image loss using negative response and image as the positive one:
\begin{equation*}
\begin{aligned}
\mathcal{L}_{\mathrm{Image-Sym}}\left(\mathcal I^+,\mathcal I^-,\mathcal A^+,\mathcal A^-;\mathcal Q\right)=\ \ \ \mathcal L_\mathrm{Image}(&\mathcal I^+,\mathcal I^-;\mathcal Q,\mathcal A^+)+\mathcal L_\mathrm{Image}(\mathcal I^-,\mathcal I^+;\mathcal Q,\mathcal A^-) \\
= - \log \sigma\bigg( 
&\beta_1 \cdot \left[ \log \frac{ \pi_\theta(\mathcal A^+ | \mathcal I^+, \mathcal Q) }{ \pi_{\mathrm{ref}}(\mathcal A^+ | \mathcal I^+, \mathcal Q) } - 
\log \frac{ \pi_\theta(\mathcal A^+ | \mathcal Q) }{ \pi_{\mathrm{ref}}(\mathcal A^+ | \mathcal Q) } \right] \\
+ &\beta_2 \cdot \left[ \log \frac{ \pi_\theta(\mathcal A^+ | \mathcal Q) }{ \pi_{\mathrm{ref}}(\mathcal A^+ | \mathcal Q) } - 
\log \frac{ \pi_\theta(\mathcal A^+ | \mathcal I^-, \mathcal Q) }{ \pi_{\mathrm{ref}}(\mathcal A^+ | \mathcal I^-, \mathcal Q) } \right] \bigg)\\
 - \log \sigma\bigg( 
&\beta_1 \cdot \left[ \log \frac{ \pi_\theta(\mathcal A^- | \mathcal I^-, \mathcal Q) }{ \pi_{\mathrm{ref}}(\mathcal A^- | \mathcal I^-, \mathcal Q) } - 
\log \frac{ \pi_\theta(\mathcal A^- | \mathcal Q) }{ \pi_{\mathrm{ref}}(\mathcal A^- | \mathcal Q) } \right] \\
+ &\beta_2 \cdot \left[ \log \frac{ \pi_\theta(\mathcal A^- | \mathcal Q) }{ \pi_{\mathrm{ref}}(\mathcal A^- | \mathcal Q) } - 
\log \frac{ \pi_\theta(\mathcal A^- | \mathcal I^+, \mathcal Q) }{ \pi_{\mathrm{ref}}(\mathcal A^- | \mathcal I^+, \mathcal Q) } \right] \bigg)
\end{aligned}
\end{equation*}

Anchor loss~$\mathcal L_\mathrm{Anchor}$ directly enforces the probability of positive response to be higher for intuitively better optimization results.
\begin{equation*}
    \mathcal L_\mathrm{Anchor}(\mathcal A^+,\mathcal A^-; \mathcal Q,\mathcal I^+)=-\log\sigma\bigg(\beta\cdot \log \frac{ \pi_\theta(\mathcal A^+ | \mathcal I^+, \mathcal Q) }{ \pi_{\mathrm{ref}}(\mathcal A^+ | \mathcal I^+, \mathcal Q) } \bigg)
\end{equation*}

Bi-directional anchor loss~$\mathcal L_\mathrm{Bi-Anchor}$ not only exerts supervision on the positive response, but it also makes the negative response probability to be lower.
\begin{equation*}
    \mathcal L_\mathrm{Bi-Anchor}(\mathcal A^+,\mathcal A^-; \mathcal Q,\mathcal I^+)=-\log\sigma\bigg(\beta\cdot \log \frac{ \pi_\theta(\mathcal A^+ | \mathcal I^+, \mathcal Q) }{ \pi_{\mathrm{ref}}(\mathcal A^+ | \mathcal I^+, \mathcal Q) } \bigg) +\log\sigma\bigg( \beta\cdot\log \frac{ \pi_\theta(\mathcal A^- | \mathcal Q) }{ \pi_{\mathrm{ref}}(\mathcal A^- | \mathcal Q) }\bigg)
\end{equation*}

\input{appendix/algorithm}

\subsection{Settings}\label{app.sub.settings}
By default, we use the following settings:

\paragraph{Software infrastructure.}
In our implementation, we deploy the non-training LLM and diffusion models as services using FastAPI. During training, the system interacts with these services via API calls to obtain feedback, image prompts, and the paths to generated images.

\paragraph{Models.}
The LLM we use for judging response and providing image-generation prompt is \texttt{Qwen-2.5-7b-instruct} (https://huggingface.co/Qwen/Qwen2.5-7B-Instruct). The diffusion model for image generation is \texttt{FLUX.1-dev} (https://huggingface.co/black-forest-labs/FLUX.1-dev). 

\paragraph{Training}
Both the 7B and 13B models are trained for a single epoch using a cosine learning rate schedule with a global batch size of 16. 
We set $\beta=\beta_1=\beta_2=0.1$ in Eq.~\ref{eq:text-dpo} and Eq.~\ref{eq:image-dpo}. Learning rates are 1e-6 for 7B model and 5e-7 for 13B model.

\paragraph{Sampling and Filter.}
The score is between 0 and 10, which 10 means a perfect response and 0 means a totally incorrect response. We sample 16 responses for one query and set the lower-bound margin $\delta$ to 3. Moreover, the quality criterion coefficients $\tau_\mathrm{pos}=\tau_\mathrm{neg}=5$, which means the score of positive response should be at least 6 and negative response be at most 4. The \textbf{temperature} of the LLM scorer is 0.1.

Our hyperparameter settings are based on preliminary experiments and empirical intuition. We observe that when the model assigns a score between 0 and 3, the responses tend to contain significant errors, while scores of 7 and above generally indicate correct answers. The more strict the preference filtering criteria is, the higher the data quality tends to be; however, this also leads to fewer preference pairs satisfying the condition. Therefore, our choice of hyperparameters is based on a balance among empirical observations, data quantity, and data quality.
\paragraph{Image Generation.} 
For image prompt generation, we set the model's \textbf{temperature} as 0.1, \textbf{top\_p} as 0.9, and \textbf{max\_new\_tokens} as 128. We generate a $384\times384$ image given the prompt with \textbf{num\_inference\_steps}=40 and \textbf{guidance\_scale}=7.5.

We perform ablation and further study using LLaVA-1.5-7B. The following describes the relevant experimental settings.

\subsubsection{Ablation on Loss Functions}
We fine-tune the model for one epoch using data generated by the model itself immediately before training, following the OViP data construction pipeline.

For \textit{iterative training}, we first fine-tune the base model on the original dataset using DPO to obtain a stronger initialization. We then sample and filter 4,730 instances as the second-stage contrastive dataset, which remains fixed across all variants. To improve supervision quality, model responses are annotated using \texttt{DeepSeek-V3} for more accurate reward estimation.

\subsubsection{Ablation on Online Learning}
Although online methods can continuously improve when trained with another epoch, we conduct the experiment with one epoch for both online and offline methods.

\subsection{Results}\label{app.sub.OOresult}
The training results for online and offline are shown in \autoref{tab:app.abl.detailOO}. Online training is significantly more effective in mitigating hallucinations.

\begin{table}[t]
\centering
\scriptsize
\caption{Online v.s. Offline detailed results}
\begin{tabular}{@{}lccccccccc@{}}
\toprule
& \multicolumn{3}{c}{AMB$_{\mathrm{gen}}$} & MMHal & \multicolumn{2}{c}{ObjectHal} & LV     & AMB$_{\mathrm{dis}}$  &
    \\ 
& Chair$\downarrow$      & Cover$\uparrow$        & \cellcolor{gray!30}\textbf{F1}$\uparrow$           & \cellcolor{gray!30}\textbf{Score}$\uparrow$ & Chair$_r\downarrow$        & \cellcolor{gray!30}\textbf{F1}$\uparrow$          & \cellcolor{gray!30}\textbf{Score}$\uparrow$        & \cellcolor{gray!30}\textbf{F1}$\uparrow$     &    \\
\midrule

Baseline& 7.1 & 50.0 & 65.01 & 1.90  & 51.38 & 72.40       &57.20&85.5&\\
\midrule
DPO$_\mathrm{online}$&\textbf{4.7}&\textbf{50.0}&\textbf{65.59}&\textbf{2.38}&\textbf{31.58}&71.70&\textbf{56.10}&\textbf{86.7}&\\
DPO$_\mathrm{offline}$&7.0&48.3&63.58&2.06&52.61&\textbf{72.55}&53.60&85.9&\\
\midrule
OViP$_\mathrm{online}$&\textbf{4.0}&\textbf{51.1}&\textbf{66.70}&\textbf{2.52}&\textbf{33.22}&\textbf{73.50}&\textbf{63.10}&\textbf{87.1}&\\
OViP$_\mathrm{offline}$&5.2&49.9&65.38&2.35&46.34&72.39&60.20&86.6&\\
\bottomrule
\end{tabular}\label{tab:app.abl.detailOO}
\end{table}

\subsection{Training Dynamics}\label{app:sub.training_dynamics}
\begin{figure}
    \centering
    \begin{minipage}{0.45\linewidth}
        \includegraphics[width=0.9\linewidth]{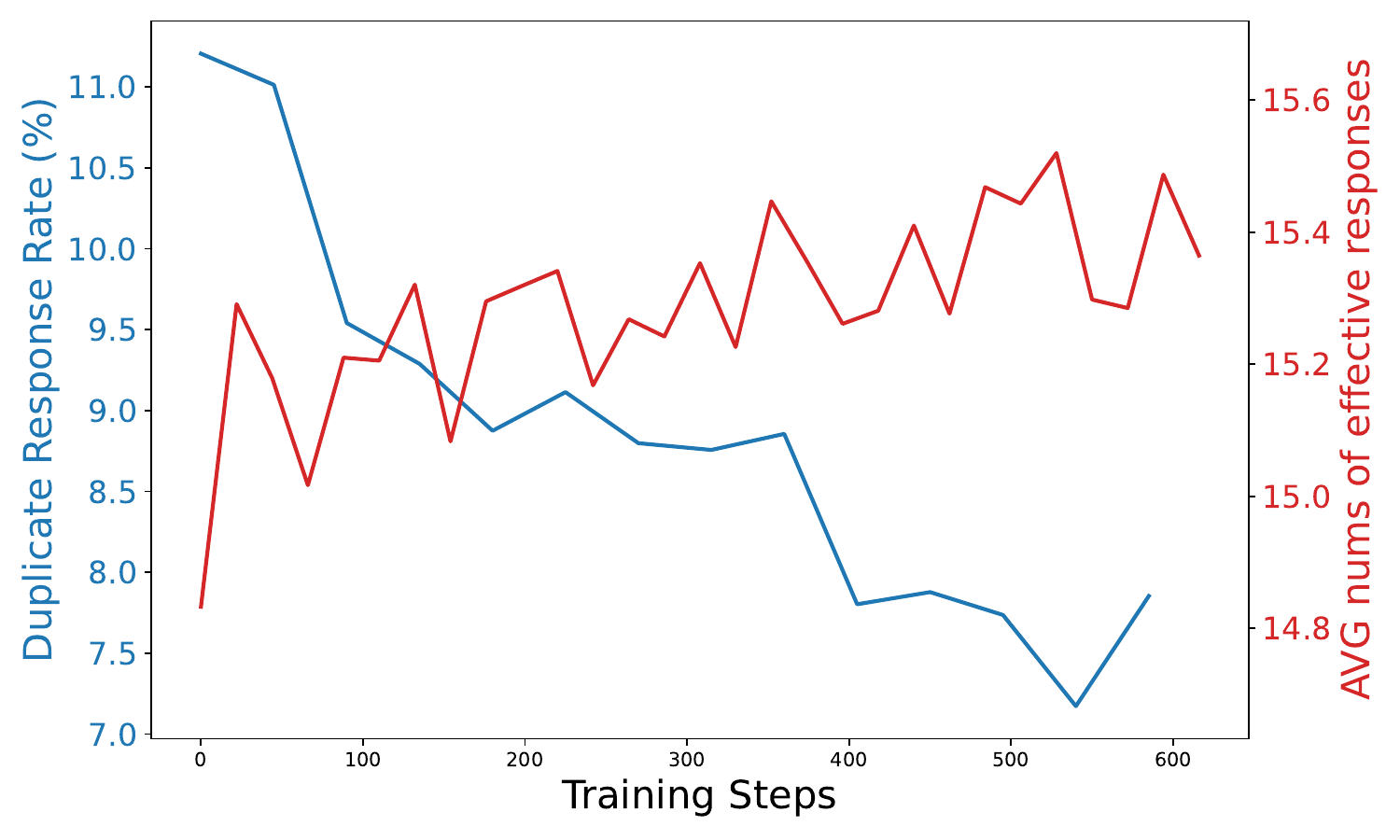}
    \caption{Sampling statistics during training. The blue curve shows the probability that, when sampling 16 responses with Temperature 1.2 for the same prompt, multiple identical responses appear (i.e., the number of distinct responses is fewer than 14). The red curve shows the average number of distinct responses obtained when sampling 16 times with Temperature 1.2.}
    \label{fig:app.sub.duplicate_response}
    \end{minipage}
    \hfill
    \begin{minipage}{0.45\linewidth}
        \includegraphics[width=0.9\linewidth]{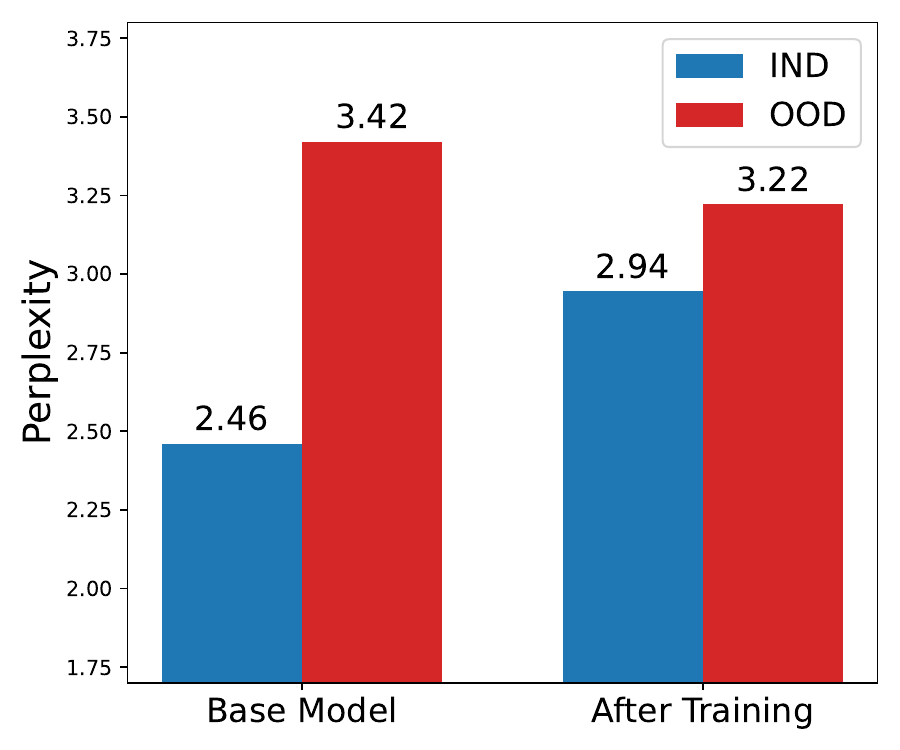}
\caption{Perplexity changes for IND and OOD sequences. The perplexity of base model's generation is relatively low.}
        \label{fig:app.sub.ppl_barplot}
    \end{minipage}
\end{figure}

The model’s initially skewed output distribution leads to a high rate of duplicate samples~(in \autoref{fig:app.sub.duplicate_response}, Duplicate Response Rate surpasses 11.0\% at first) and very low perplexity~\autoref{fig:app.sub.ppl_barplot} in the generated responses, which is not conducive to optimization. In the early stages of training, the output distribution gradually flattens, but the limited exploratory scope prevents the model from identifying the correct optimization direction, resulting in stagnation of performance metrics. As the coverage of the distribution expands, the model can effectively explore the correct update directions, allowing training to get on track and performance to accelerate.

%% file: appendix/algorithm.tex
\begin{table}[t]
\centering
\renewcommand{\arraystretch}{1.2}
\caption{OViP pseudocode}
\label{tab:app:algorithm}
\begin{tabular}{@{}p{0.95\linewidth}@{}}
\toprule
\textbf{Algorithm 1} Algorithm of \ovip{} \\
\midrule
\textbf{Input:} training dataset $\mathcal{D} = \{(\mathcal{I}^+, \mathcal{Q}, \mathcal{A}^*)\}$; \\
\hspace{2.5em} target model $\pi$; reward model $\mathrm{G}_{\mathrm{r}}$; prompt generator $\mathrm{G}_{\mathrm{diff}}$; diffusion model $\mathrm{diff}$ \\

\textbf{Initialize:} experience buffer $\mathcal{B} \leftarrow \emptyset$ \\
\textbf{Output:} optimized model $\pi$ \\

\textbf{for} each $(\mathcal{I}^+, \mathcal{Q}, \mathcal{A}^*) \in \mathcal{D}$ \textbf{do} \\
\quad Sample candidate responses $\{\mathcal{A}^i\}_{i=1}^k \sim \pi(\cdot|\mathcal{I}^+, \mathcal{Q})$ \\
\quad Compute reward scores: $r^i = \mathrm{G}_{\mathrm{r}}(\mathcal{A}^i, \mathcal{A}^*)$ \\
\quad Compute standard deviation $\sigma_r$ of $\{r^i\}$ \\

\quad Initialize temporary pair list $\mathcal{T} \leftarrow \emptyset$ \\

\quad \textbf{while} $\exists$ $(\mathcal{A}^+, \mathcal{A}^-)$ satisfying: \\
\quad \quad $|r^+ - r^-| > \max(\delta, 2\sigma_r)$, $r^+ > \tau_{\mathrm{pos}},\ r^- < \tau_{\mathrm{neg}}$ \textbf{do} \\
\quad \quad Add $(\mathcal{A}^+, \mathcal{A}^-)$ to $\mathcal{T}$ and remove from candidate pool \\
\quad \textbf{end while} \\

\quad \textbf{if} $\mathcal{T} = \emptyset$ and $\min_i r^i < \tau_{\mathrm{neg}}$ \textbf{then} \\
\quad \quad Let $\mathcal{A}^-$ be the lowest-scoring response \\
\quad \quad Add $(\mathcal{A}^*, \mathcal{A}^-)$ to $\mathcal{T}$ \\
\quad \textbf{endif} \\

\quad \textbf{for each} $(\mathcal{A}^+, \mathcal{A}^-) \in \mathcal{T}$ \textbf{do} \\
\quad \quad Generate prompt: $\mathcal{T}^- = \mathrm{G}_{\mathrm{diff}}(\mathcal{A}^+, \mathcal{A}^-)$ \\
\quad \quad Synthesize image: $\mathcal{I}^- = \mathrm{diff}(\mathcal{T}^-)$ \\
\quad \quad Add $(\mathcal{I}^+, \mathcal{I}^-, \mathcal{Q}, \mathcal{A}^+, \mathcal{A}^-)$ to buffer $\mathcal{B}$ \\
\quad \textbf{end for} \\

\quad \textbf{if} $|\mathcal{B}| \geq N$ \textbf{then} \\
\quad \quad Sample $N$ samples from $\mathcal{B}$ for training \\
\quad \quad Compute total loss: $\mathcal{L}_{\mathrm{OViP}}$\\
\quad \quad Update $\pi \leftarrow \pi - \eta \nabla_\pi \mathcal{L}_{\mathrm{OViP}}$ \\
\quad \textbf{endif} \\
\textbf{end for} \\
\bottomrule
\end{tabular}

\end{table}

%% file: appendix/efficiency.tex
\begin{figure}
    \centering
    \includegraphics[width=0.5\linewidth]{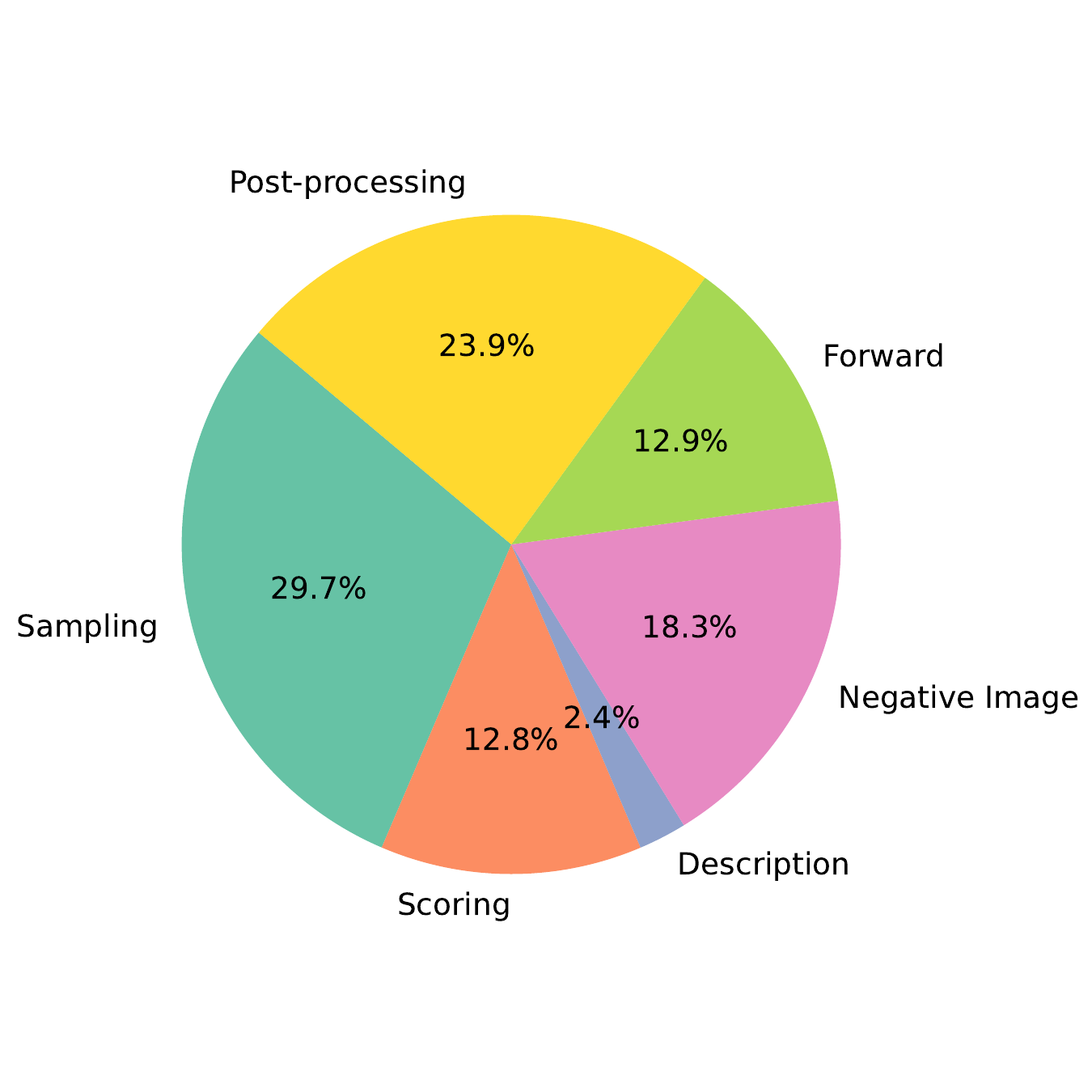}
    \caption{Time consumption for each stage during training.}
    \label{fig.app:time-consuming}
\end{figure}

OViP training takes approximately 17 hours on 7× A800 (40G) GPUs. Among them, 4 GPUs are allocated for VLM training, 1 GPU for LLM deployment, and 2 GPUs for diffusion model deployment. We divide each training step into six stages: sampling (response generation), scoring (response evaluation), description (image prompt construction), negative image (counterfactual image generation), forward (model inference), and post-processing. \autoref{fig.app:time-consuming} illustrates the proportion of time spent on each stage, where post-processing refers to the period after forward propagation and before the next training step begins, including gradient accumulation, backpropagation, optimizer updates, and other related operations.

Excluding post-processing, the most time-consuming component is the sampling stage, similar to reinforcement learning. This is because it requires autoregressive generation of 16 responses, one token at a time. The second most expensive stage is negative image generation. To reduce latency, we parallelize this process by assigning two diffusion models to handle image generation requests from four sampling subprocesses.

Additionally, since the experience buffer is implemented independently in our system, repeated sampling by one subprocess may block others due to synchronization constraints. This can indirectly slow down the forward and post-processing stages as some processes await completion.

%% file: appendix/limitations.tex
This work introduces an online training framework that integrates dual contrastive learning across vision and language. While our loss function follows the DPO formulation, we do not explore how existing reinforcement learning algorithms—such as PPO or GRPO—could be effectively combined with image-level contrastive objectives. In terms of evaluation, although we identify and discuss several limitations of prior protocols and propose improved metrics and procedures, the current benchmarks still fall short of fully capturing model capability. We manually identified a subset of erroneous cases through inspection, but did not conduct a comprehensive correction. Lastly, our data filtering strategy during sampling has not been carefully tuned, and a more refined design could potentially lead to better training dynamics and model performance.

%% file: appendix/prompt.tex
\begin{table*}[h]
\centering
    \begin{tcolorbox}
[colback=black!5!white,colframe=gray!15!gray,width=0.9\textwidth,,title={Prompt for Quality Judgment}]
\small
\# Task\\
Your role is as a discerning assistant tasked with evaluating model responses for multimodal tasks (though you have no access with the image). Upon being presented with a question that requires the interpretation of both text and images, you will receive two distinct responses. The first is crafted by our sophisticated multimodal model, while the second represents an approximate ideal answer--it may be incomplete. Your objective is to meticulously and precisely assess the model-generated response (the former) based on the provided reference answer (the latter).\\

- Here's how you should approach the assessment process:\\
\phantom{1111}1. The quality of the response depends on its accuracy and the degree of adherence to the correct answer. Therefore, if the response is much more detailed than the reference answer, it should not be considered a very good response (although it may still be considered a good one).\\
\phantom{1111}2. Directly provide the score of the response, with a full score of 10. Your response should follow this format: "Score: [x]\textbackslash n", where "[x]" represents the score you give, and "\textbackslash n" is a line break.\\
\phantom{1111}3. Please do not provide additional reasoning, just give the score directly.\\

\# Question\\
\{question\}\\

\# Response\\
\{response\}\\

\# Correct answer\\
\{answer\}
 \end{tcolorbox}
 \caption{The prompt for judging model response given the answer.}
 \label{app:tab:prompt for judgment}
\end{table*}

\begin{table*}
\centering
    \begin{tcolorbox}
[colback=black!5!white,colframe=gray!15!gray,width=0.9\textwidth,,title={Prompt for image generation-Part 1}]
\small
\# Task  \\
Given an unknown image-related question, a correct answer, and an inaccurate response, carefully analyze the differences between the response and the answer. Then, provide a brief description of the image so that it aligns with the correct answer and differs from the incorrect responses. In other words, infer the content of the image.\\

\# Example

[Example 1] \\
**<Question>**  \\
What is on the sandwich?  \\

**<Answer>**\\
The sandwich has tomatoes and lettuce on it.  \\

**<Response>**  \\
The sandwich has a slice of egg and tomato on it.\\

**<Output Description>**\\
A sandwich with only tomatoes and lettuce on it.\\

**Explanation**: The answer mentions lettuce and tomato, while the incorrect response mentions tomato and egg. So there is no egg on the sandwich.\\

[Example 2]\\
**<Question>**  \\
Can you point out the details that make this image unique?  \\

**<Answer>**\\
In the image, there is a plate with a slice of pizza topped with tomatoes, herbs, and cheese. The distinctive detail about the image is that the pizza is missing two slices, leaving just one slice remaining on the plate. This suggests that someone has already started enjoying the pizza. The slice appears to be well-cooked and freshly served, creating an appetizing and mouthwatering scene for the viewer.  \\

**<Response>**  \\
In the image, there is a slice of pizza on a plate with tomatoes and cheese. The pizza appears to be homemade and has been cut into two pieces. The tomatoes are sliced in half, revealing their juicy interior. The cheese on top of the pizza is melted, creating a delicious-looking dish. Additionally, there is a fork nearby, suggesting that someone might be planning to enjoy this pizza soon.  \\

**<Output Description>**\\
A plate with a one-third remaining piece of pizza, topped with herbs, cheese, and tomatoes; someone has finished eating and left.\\

**Explanation**: The answer mentions that only one-third of the pizza remains and that someone has just finished eating and left, which is inconsistent with the response. Therefore, the image should include these two features.\\

 \end{tcolorbox}
 \caption{The prompt for image generation.}
 \label{app:tab:prompt for imageprompt distort1}
\end{table*}

\begin{table*}
\centering
    \begin{tcolorbox}
[colback=black!5!white,colframe=gray!15!gray,width=0.9\textwidth,,title={Prompt for image generation-Part 2}]
\small
[Example 3]\\
**<Question>**\\
Bird or cow?\\

**<Answer>**\\
Bird\\

**<Response>**\\
The bird in the image is a small, brown and white bird with a distinctive head shape and coloration. It is not a cow. The bird is perched on a branch, which is situated in front of a white building.\\

**<Output Description>**\\
A big, blue bird perched on a branch in front of a black building.\\

**Explanation**: Both the answer and the response mention the bird, but the response is more detailed. So the description should be contrastive to the features of the bird in the response.\\

\# Requirements\\
- The description should be brief but precise.  \\
- If both the answer and the response are long, focus on describing the one or two most significant differences.\\
- Do not provide any analysis or explanation; only describe the image.\\
- A common approach is to describe what is present in the image and what is missing.\\

**<Question>**\\
{question}\\
**<Answer>**\\
{answer}\\
**<Response>**\\
{response}\\

**<Output Description>**\\
 \end{tcolorbox}
 \caption{The prompt for image generation.}
 \label{app:tab:prompt for imageprompt distort2}
\end{table*}